\PassOptionsToPackage{numbers, compress}{natbib}
\documentclass[11pt,a4paper,logo]{googledeepmind}

\setleftlogo[180pt]{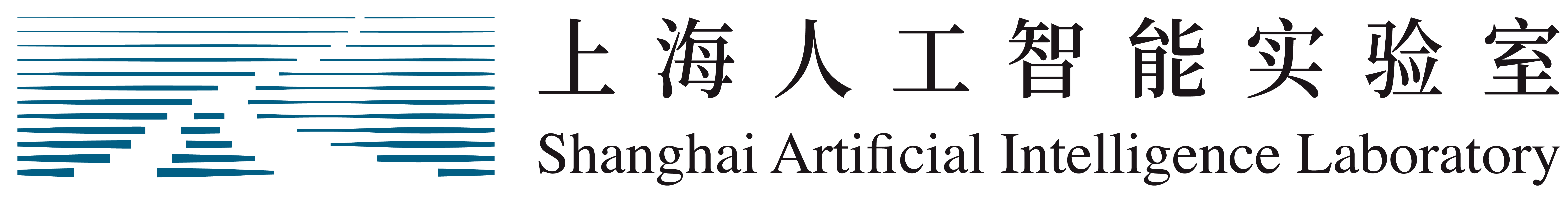}
\setrightlogo[25pt]{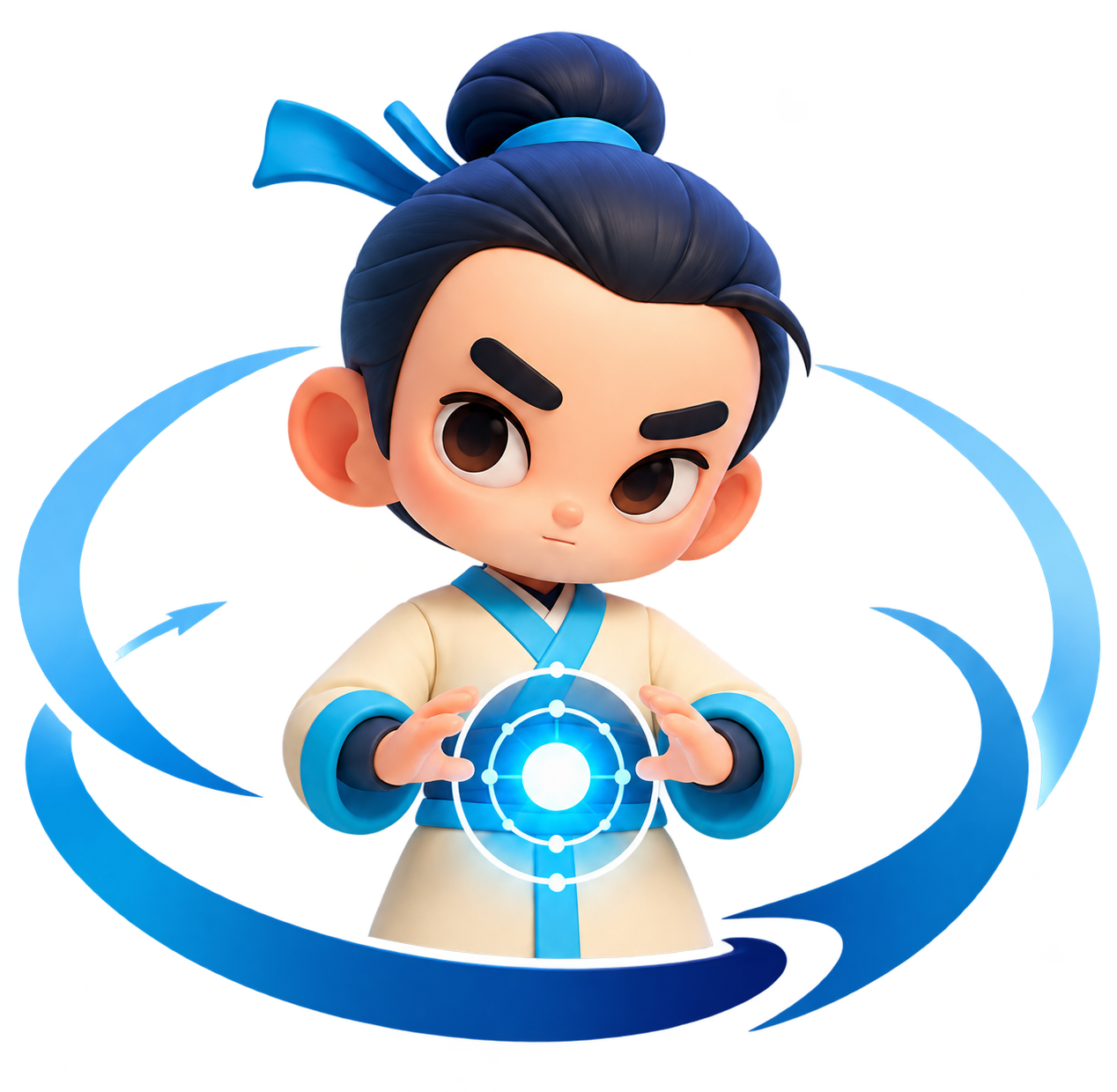}

\usepackage{natbib}
\setcitestyle{numbers,square,comma}
\hypersetup{
    colorlinks=true,
    citecolor=blue,
    linkcolor=blue,
    urlcolor=blue
}

\usepackage{url}
\usepackage{booktabs}
\usepackage{amsfonts}
\usepackage{amsmath}
\usepackage{amssymb}
\usepackage{mathtools}
\usepackage{bm}
\usepackage{xcolor}
\usepackage{enumitem}
\usepackage{multirow}
\usepackage{wrapfig}
\usepackage{colortbl}
\usepackage{tabularx}
\usepackage{array}
\usepackage{graphicx}
\usepackage[capitalize,noabbrev]{cleveref}
\usepackage[most,skins,theorems]{tcolorbox}
\usepackage{tikz}

\usepackage{arydshln}
\expandafter\let\csname endtabular*\endcsname\endtabular
\usetikzlibrary{shapes.geometric}

\theoremstyle{plain}

\theoremstyle{definition}

\theoremstyle{remark}

\definecolor{warm}{HTML}{C45A5A}
\definecolor{cool}{HTML}{377EB8}
\definecolor{mygreen}{HTML}{298D66}
\definecolor{myred}{HTML}{F94153}
\definecolor{tts_or}{HTML}{8AD879}
\definecolor{tts_sc}{HTML}{53D2DC}
\definecolor{srpo}{HTML}{4DBBD5}
\definecolor{octopus}{HTML}{669999}

\newcolumntype{L}{>{\hspace{2.2pt}}l<{\hspace{2.2pt}}}
\newcolumntype{Y}{>{\hspace{2.2pt}\centering\arraybackslash}X<{\hspace{2.2pt}}}
\newcolumntype{P}[1]{>{\centering\arraybackslash}p{#1}}
\newcounter{tablepanel}

\newcommand{\aout}[1][]{A_{\text{out}\ifx\relax#1\relax\else,#1\fi}}
\newcommand{\aproc}[1][]{A_{\text{proc}\ifx\relax#1\relax\else,#1\fi}}
\newcommand{\atotal}[1][]{A_{\text{total}\ifx\relax#1\relax\else,#1\fi}}

\usepackage{caption}

\newcommand{\Projectname}{\textsc{Janus-MOPD}}
\newsavebox{\safintitlebox}
\title{%
    \sbox{\safintitlebox}{%
    \begin{minipage}[c]{0.90\linewidth}
        \raggedright
        \fontsize{17}{20}\selectfont
        \Projectname{}: Tracing the Student Learning\\
        Dynamics in Multi-Teacher On-Policy Distillation
    \end{minipage}}%
    \raisebox{-\dp\safintitlebox}{%
        \includegraphics[height=\dimexpr\ht\safintitlebox+\dp\safintitlebox\relax]{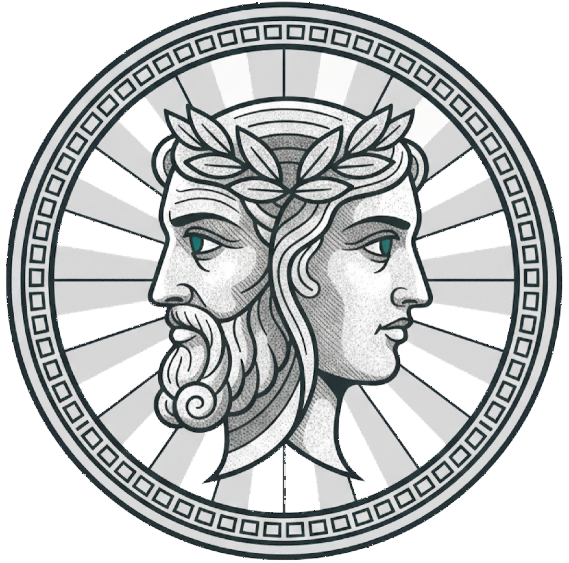}%
    }\hfill%
    \usebox{\safintitlebox}%
}

\author[1,2]{Zongji Yu}
\author[1,3]{Ming Zhang}
\author[1,2]{Dongcheng Zhang}
\author[1]{Jiawei Zhou}
\author[1,2]{Haotian Pang}
\author[2]{Yiliu Sun}

\author[1,4]{Xiang Wang}
\author[1,3$\dag$]{Youbang Sun}
\author[1,$\dag$]{Chaochao Lu}
\author[2,$\dag$]{Chen Gong}
\affil[1]{Shanghai AI Laboratory}
\affil[2]{Shanghai Jiao Tong University}
\affil[3]{Tsinghua University}
\affil[4]{University of Science and Technology of China}

\renewcommand\Affilfont{\normalfont\fontsize{7.8}{9}\selectfont}
\makeatletter
\@namedef{@sep7}{\protect\\}
\makeatother
\correspondingauthor{$^\dag$ Co-corresponding authors}

\definecolor{AbstractBgColor}{HTML}{F4F7FB}
\usepackage{etoolbox}
\patchcmd{\maketitle}{\vskip20pt}{\vskip0pt}{}{}

\makeatletter
\renewcommand{\abscontent}{%
  \begingroup
    \setlength{\parskip}{0pt}%
    \absfont
    \vspace{3pt}
    {\color{black}\hrule height 0.6pt}%
    \vspace{6pt}
    \noindent\textbf{Abstract}\normalfont~$|$~\theabstract\par
    \vspace{2pt}
  \endgroup
}
\makeatother

\usepackage{tocloft}

\hypersetup{
    pdftitle={Janus-MOPD: Tracing the Student Learning Dynamics in Multi-Teacher On-Policy Distillation},
    pdfauthor={Zongji Yu, Ming Zhang, Dongcheng Zhang, Youbang Sun, Chaochao Lu, Chen Gong},
    pdfsubject={Multi-teacher on-policy distillation and student learning dynamics}
}
\begin{abstract}
Multi-domain capabilities are critical for practically applicable Large Reasoning Models (LRMs).
Therefore, Multi-teacher On-Policy Distillation (MOPD) has become a crucial part of the post-training process, integrating the capabilities of multiple domain-specific teacher models into a single student model.
However, the student usually fails to acquire multi-domain capabilities simultaneously due to undesirable domain heterogeneity.
Existing works have address this problem in view of temporal uneven efficiency and spatial hard
allocation in external multi-teacher supervision. While in broader multi-domain training, varying degrees of performance degradation is still largely unexplained. 
In this paper, we shift the focus on how the student learns and integrates multi-domain capabilities internally, and then propose \textbf{\textsc{Janus-MOPD}}, a general learning dynamics framework of geometric representations for LRMs.
Specifically, our main focus is on the internal performance of multi-domain capabilities on both temporal and spatial dimensions within the student.
Temporally, our study shows that the student exhibits a Expansion-Consolidation-Optimization (ECO) three-phase learning dynamics, together with inherent limitations in learning efficiency and scaling.
Spatially, we summarize two types of subspace locking within and across domains by tracing the representation space of the student, while explaining its clear learning directions but inherent cross-domain conflict.
Therefore, \textsc{Janus-MOPD} yields a unified learning dynamics formulation, a geometric tracing methodology, and systematic findings of how the student learns and integrates multi-domain capabilities.
Overall, we provide a \textsc{Janus} perspective to internally trace the continual evolution of the multi-domain capabilities within the model itself.

\end{abstract}

\begin{document}
\vspace*{-15pt}
\begin{tcolorbox}[
    colback=AbstractBgColor,colframe=AbstractBgColor,
    arc=5pt,boxrule=0pt,left=8pt,right=8pt,
    top=6pt,bottom=0pt,parbox=false,width=\textwidth,
    before skip=0pt,after skip=8pt]
\maketitle
\end{tcolorbox}
\begin{quote}
\vspace{-2mm}
\textit{``Every door has two fronts, this way and that.''}
\hfill--- Ovid, Fasti, I.135
\vspace{-1mm}
\end{quote}

\begin{minipage}{\textwidth}

  \centering
  \centering
  \fontsize{8pt}{9pt}\selectfont
  \setlength{\tabcolsep}{3pt}
  \renewcommand{\arraystretch}{1.12}
  \centering
  \includegraphics[width=1.0\textwidth]{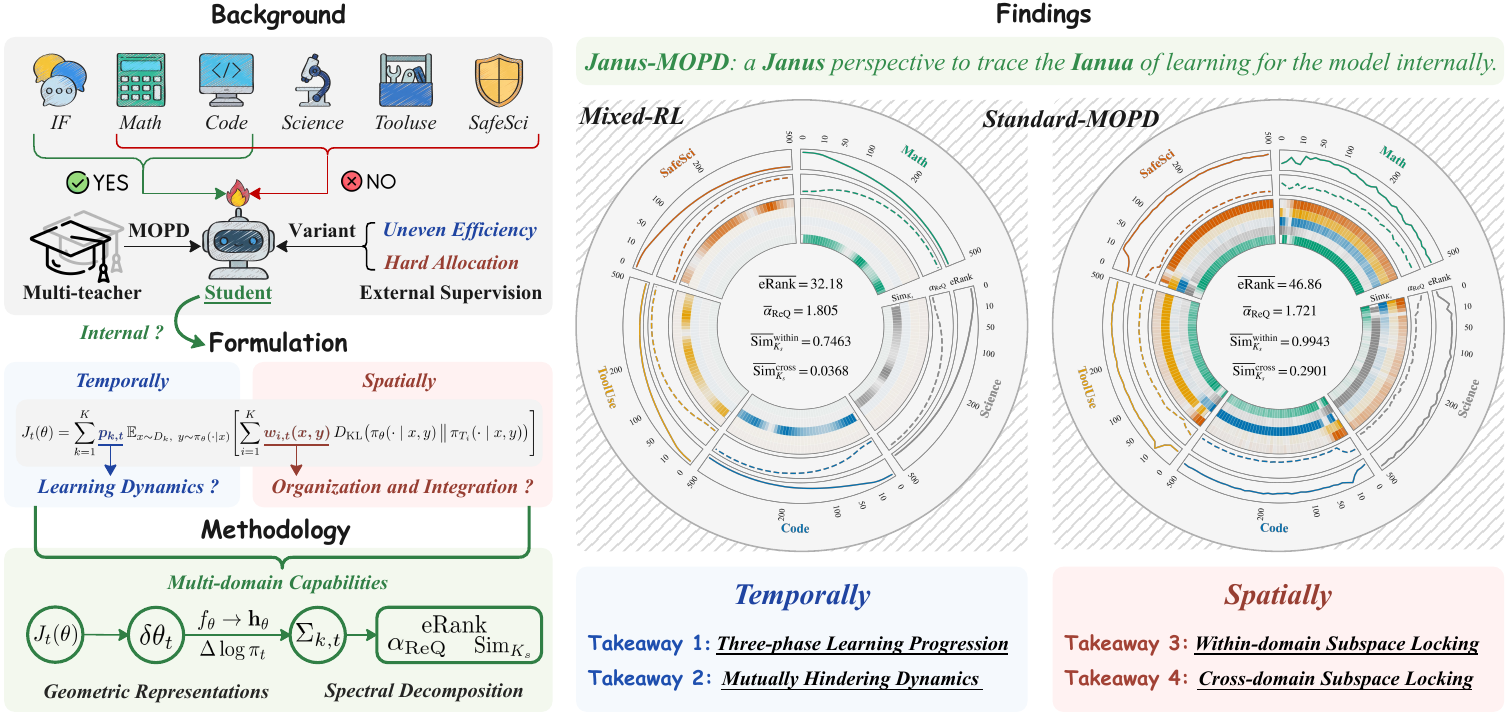}
  \captionsetup{font=small}
  \captionof{figure}{\textbf{Overview of \textsc{Janus-MOPD}.}
  \textbf{Left:} A unified formulation and geometric tracing methodology characterize student learning from temporal and spatial perspectives.
  \textbf{Right:} Mixed-RL and Standard-MOPD illustrate representative geometric traces across five domains.
  Temporal analyses reveal three-phase learning progression with mutually hindering, while spatial analyses reveal subspace locking within and across domains.}
  \label{fig:Intro}
\end{minipage}

\clearpage


\newtcolorbox{rqbox}{
  colback=Gray!5!white,
  colframe=Gray!80!black,
  size=small,
  boxed title style={empty},
  title=,
}

\section{Introduction}\label{sec:intro}

Multi-domain capabilities are critical for practically applicable Large Reasoning Models (LRMs) \citep{deepseekr1, mimov2flash2026, deepseekv42026}.
Therefore, Multi-teacher On-Policy Distillation (MOPD) \citep{mimov2flash2026,deepseekv42026,mopd2026} was proposed as a crucial part of the post-training process to integrate multi-domain capabilities into one student model.
Specifically, MOPD first trains several teachers separately on domain-specific data \citep{agarwal2024onpolicy, gu2024minillm, lu2025onpolicy}.
Subsequently, within a unified training process, it applies a domain router to match the responses generated by the student to the corresponding teacher for on-policy distillation (OPD) \citep{deepseekv42026, mopd2026}.
In practice, however, due to the domain heterogeneity, capabilities from different domains mutually hinder one another \citep{openmopd2026, roadmopd2026, immopd2026, deltamopd2026}.
This trade-off makes it difficult for the student to acquire multi-domain capabilities simultaneously.

Existing works attribute this phenomenon to uneven efficiency or hard allocation in external multi-teacher supervision. 
Respectively, they propose to coordinate teacher-student gaps across domains\citep{openmopd2026,d3mopd2026} and design soft router for multi-teacher joint supervision \citep{roadmopd2026,mopdrouter2026,trustmopd2026, camopd2026, dnmopd2026}.
We observe that these methods focus mainly on addressing temporal imbalance and spatial one-sidedness in external multi-teacher supervision.
However, in broader multi-domain training, varying degrees of performance degradation is still largely unexplained. 
Within this context, we claim that studying these problems requires shifting the focus from external multi-teacher supervision to how the student learns and integrates multi-domain capabilities internally.
Accordingly, we instead propose to trace the student learning dynamics  \citep{ren2025learning, li2025tracing} from both temporal and spatial perspectives, and systematically study the internal performance and evolution of multi-domain capabilities within the student during the MOPD training process. 
Specifically, we empirically study MOPD through two questions:

\begin{rqbox}
\begin{itemize}
\item {\textbf{Temporally}: what are the learning dynamics of the student for multi-domain capabilities?}
\item {\textbf{Spatially}: how does the student organize and integrate multi-domain capabilities?}
\end{itemize}
\end{rqbox}



To answer these questions, we propose \textbf{\textsc{Janus-MOPD}}, which uses spectral analysis \citep{jolliffe2002principal,golub2013matrix,bjorck1973angles} to quantify the evolution of representation geometry within the student across MOPD paradigms.
Firstly, \textsc{Janus-MOPD} focuses on the spectral properties of covariance matrices computed from last-token representations, and  characterizes the representation geometry by tracing the evolution of three metrics derived from their eigenspectra \citep{li2025tracing, garrido2023rankme, agrawal2022areq}.
Based on this formulation, we introduce a tracing methodology to analyze the internal performance and evolution within the student from complementary temporal and spatial perspectives.
Temporally, we study the learning progression between MOPD and other post-training paradigms \citep{ren2025learning, li2026dualnature,zheng2026resi}, along with the scaling dynamics of domain representations within the student. 
Spatially, we analyze the organization and integration of multi-domain capabilities, along with their underlying interactions during the training process.

Our extensive experiments show that the student internally exhibits a phase-wise learning dynamics, capability trade-offs, and locking properties within and across domains.
Temporally, compared with Mixed-RL \citep{deepseekr1,yu2026mcpo}, the student in MOPD exhibits learning dynamics more akin to those of “students.”. 
Specifically, during the learning of multi-domain capabilities, the student clearly exhibits three-phase learning progression of Expansion-Consolidation-Optimization (ECO) with domain heterogeneity.
Moreover, inherent learning limitations in efficiency and scaling cause capabilities across domains to interfere.
Spatially, we identify two types of subspace locking by tracing the representation space of the student. 
Within each domain, the student rapidly identifies a locking low-rank subspace and concentrates its learning direction within it.
While across domains, the student also quickly locks cross-domain interactions into low-rank subspaces. 
Anyway, these distinct and locked learning dynamics provide clear learning directions, but also induce inherent cross-domain conflicts.

Based on \textbf{\textsc{Janus-MOPD}}, our main contributions are summarized as follows:
\vspace{-3mm}
\begin{itemize}
 \item \textbf{Formulation}. We formulate a common learning dynamics framework to characterize the internal performance and evolution of multi-domain capabilities within the student.
 \vspace{1mm}
 \item \textbf{Methodology}. We introduce a tracing methodology to compare how geometric representations evolve under existing MOPD methods from complementary temporal and spatial perspectives.
 \vspace{1mm}
 \item \textbf{Findings}. We reveal a series of takeaways from extensive experiments on how the student organizes and integrates multi-domain capabilities internally.
\end{itemize}
\vspace{-3mm}
Overall, \textsc{Janus-MOPD} yields a unified learning dynamics formulation, a geometric tracing methodology, and systematic findings of how the student learns multi-domain capabilities,  providing a \textsc{Janus} perspective to internally trace the continual evolution of the multi-domain capabilities.
\section{Problem Formulation and Methodology}
\label{sec:prelim}

In this section, we formulate a common learning dynamics framework to characterize the internal performance of multi-domain capabilities within the student. 
Specifically, we first cast MOPD and its variants under two supervision coefficients (Section~\ref{sec:prelim-mopd}), then define the learning dynamics of geometric representations on an observation matrix (Section~\ref{sec:prelim-geometry}), and finally introduce the tracing methodology with three metrics from complementary temporal and spatial perspectives (Section~\ref{sec:prelim-metrics}).

\subsection{Multi-Teacher On-Policy Distillation}
\label{sec:prelim-mopd}

On-policy distillation (OPD) trains a student $\pi_\theta$ on its own rollouts under teacher supervision \citep{agarwal2024onpolicy, gu2024minillm, lu2025onpolicy}. 
Multi-Teacher On-Policy Distillation (MOPD) further extends OPD to $K$ domains within a unified training process\citep{mimov2flash2026,deepseekv42026,mopd2026}.
Specifically, each domain $k$ contributes a prompt distribution $D_k$ and a domain-specific teacher $\pi_{T_k}$, then for a prompt $x\sim D_k$ the student generates $y\sim\pi_\theta(\cdot\mid x)$ on its own. 
Therefore, the standard-MOPD objective aggregates per-domain reverse-KL losses:
\begin{equation}
\label{eq:mopd-obj}
J(\theta)=\sum_{k=1}^{K}p_k\,\mathbb{E}_{x\sim D_k,\ y\sim\pi_\theta(\cdot\mid x)}\!\left[D_{\mathrm{KL}}\!\left(\pi_\theta(\cdot\mid x,y)\,\big\|\,\pi_{T_k}(\cdot\mid x,y)\right)\right],
\end{equation}
where $p_k\geq0$ with $\sum_{k}p_k=1$ is the domain mixture, which is fixed before training by the pooled data shares $p_k=N_k/\sum_iN_i$.
The reverse-KL is summed over the tokens of $y$.
Formally, at position $j$ with context $s_j=(x,y_{<j})$, the matched teacher scores the context that the student itself generated for dense on-policy supervision.
Practical variants revise Eq.~\eqref{eq:mopd-obj} along two axes of the supervision, which we cast into one generalized objective at training step $t$:
\begin{equation}
\label{eq:mopd-general}
J_t(\theta)=\sum_{k=1}^{K}p_{k,t}\,\mathbb{E}_{x\sim D_k,\ y\sim\pi_\theta(\cdot\mid x)}\!\left[\sum_{i=1}^{K}w_{i,t}(x,y)\,D_{\mathrm{KL}}\!\left(\pi_\theta(\cdot\mid x,y)\,\big\|\,\pi_{T_i}(\cdot\mid x,y)\right)\right],
\end{equation}
where the \emph{temporal coefficient} $p_{k,t}$ decides when and how much each domain contributes, and the \emph{spatial coefficient} $w_{i,t}(x,y)$ decides which teachers supervise each context with what strength. 

Based on this formulation, we categorize existing MOPD methods into temporal and spatial perspectives.
Specifically, $p_{k,t}$ may balance token budgets and teacher--student gaps across domains \citep{openmopd2026}, or track the online per-domain reverse-KL \citep{d3mopd2026}.
Complementarily, $w_{i,t}(x,y)$ may weight the teachers by token-level relevance \citep{mopdrouter2026}, reliability \citep{trustmopd2026}, guarding \citep{roadmopd2026}, or conflict-aware normalization \citep{camopd2026, dnmopd2026}. 

Overall, in our view, the two coefficients respectively answer the \emph{temporal imbalance} of convergence rates across domains and the \emph{spatial one-sidedness} of teacher supervision. 
Setting $p_{k,t}\equiv p_k$ and $w_{i,t}=\mathbf 1[i=k]$ recovers the standard hard-routing objective of Eq.~\eqref{eq:mopd-obj}.
Therefore, every existing method is an instantiation of the coefficient pair $(p_{k,t},w_{i,t})$, and both coefficients act only on the external teacher supervision. 
How the student internally organizes and integrates the supervised multi-domain capabilities is left open by construction, which is what we formulate in Section~\ref{sec:prelim-geometry}.

\subsection{Learning Dynamics within the Student}
\label{sec:prelim-geometry}

Learning dynamics describes how the learning of specific samples influences the predictions on other samples \citep{ren2025learning}.
Concretely, it is the map from a parameter update $\delta\theta_t=\theta_{t+1}-\theta_t$ to the induced change $f_{\theta_{t+1}}(x)-f_{\theta_t}(x)$. 
For MOPD, minimizing Eq.~\eqref{eq:mopd-general} by policy gradient yields the per-step update
\begin{equation}
\label{eq:update}
\delta\theta_t=\eta\sum_{k=1}^{K}p_{k,t}\,\mathbb{E}_{x\sim D_k,\ y\sim\pi_\theta(\cdot\mid x)}\!\left[\sum\nolimits_{j}\hat A_j\,\nabla_\theta\log\pi_\theta(y_j\mid s_j)\right],
\end{equation}
where the per-token distillation $\hat A_j=\log\pi_{T_k}(y_j\mid s_j)-\log\pi_\theta(y_j\mid s_j)$ is evaluated at the student states with the mixed teacher signal of Eq.~\eqref{eq:mopd-general}.
Following the decomposition of learning dynamics~\citep{ren2025learning}, we linearize around $\theta_t$. For one reverse-KL gradient contribution at a fixed updating state $x_u$, the first-order change in the log-probability vector at an observing state $x_o$ is
\begin{equation}
\label{eq:influence}
\Delta\log\pi_t(y\mid x_o)=-\eta\,J_t(x_o)\,K_t(x_o,x_u)\,G_t(x_u)+O(\|\delta\theta_t\|^2),
\end{equation}
where $z(x)$ denotes the logits of the student at state $x$.
Formally, $J_t(x_o)=I-\mathbf 1\pi_{\theta_t}^{\top}(x_o)$ is the softmax Jacobian at the observed state, which maps the logit change at $x_o$ to the log-probability change of the student.
\mbox{$K_t(x_o,x_u)=\nabla_\theta z(x_o)\,\nabla_\theta z(x_u)^{\top}$} is the neural tangent kernel built from the logit gradients, which pairs the logit gradient of the updating state $x_u$ from the student rollouts with that of the observed state $x_o$.
$G_t(x_u)
=q_v\sum_iw_{i,t}(x_u)\left[\log\frac{q_v}{\tau_{i,v}}-D_{\mathrm{KL}}(q\|\tau_i)\right]$ is the distillation direction, which aggregates $\hat A_j$ of Eq.~\eqref{eq:update} and points from the current distribution of the student to its teacher.
The three factors separate what the update can move ($J$), which states it couples ($K$), and where it pushes ($G$).

\subsection{Dynamics Metrics of Geometric Representations}
\label{sec:prelim-metrics}

Equation~\eqref{eq:influence} observes the learning dynamics at the behavior level of single states.
Subsequently, we aggregate the same first-order field over a frozen population of probes and observe it in representation space.
For an autoregressive student, the last-token representation $\mathbf h_\theta\in\mathbb R^d$ summarizes the full context and feeds the next-token readout, which is the common site to trace learning dynamics \citep{li2025tracing}.

Specifically, the decomposition of Eq.~\eqref{eq:influence} in fact applies to any differentiable observable $f_\theta(x_o)$, and one gradient step changes $f_\theta$ to first order by $\nabla_\theta f_\theta(x_o)\,\delta\theta_t+O(\|\delta\theta_t\|^2)$.
To trace the geometric representations within the student, we take $\mathbf h_\theta=f_\theta$, in other words the predictions and representations are two readouts of the same update.
Therefore, the representation Jacobian $\nabla_\theta\mathbf h_t(x)$ replaces $J_t$ and decides what the update can move in the feature space, the kernel $K_t$ now couples every probe with the updating rollouts through \mbox{$\nabla_\theta\mathbf h_t(x_o)\,\nabla_\theta z(x_u)^{\top}$}, and $G_t$ still supplies where the update pushes.

The accumulated changes of $\mathbf h_t$ over the probes therefore record exactly the influence field of Eq.~\eqref{eq:influence}, observed where the capabilities are stored rather than where they are scored.
Subsequently,  for each domain $k$ and each step $t$, we freeze $M$ probes $\{x_{k,m}\}_{m=1}^{M}$ and collect the second moment of the centered representations into the observation matrix
\begin{equation}
\label{eq:feature-covariance}
\Sigma_{k,t}=\frac{1}{M}\sum_{m=1}^{M}\big(\mathbf h_t(x_{k,m})-\bar{\mathbf h}_{k,t}\big)\big(\mathbf h_t(x_{k,m})-\bar{\mathbf h}_{k,t}\big)^{\!\top},
\end{equation}
where $\bar{\mathbf h}_{k,t}=\frac{1}{M}\sum_{m=1}^{M}\mathbf h_t(x_{k,m})$ is the mean representation over the probes.
Writing its eigendecomposition $\Sigma_{k,t}=\sum_{a=1}^{d}\sigma_a\mathbf u_a\mathbf u_a^{\top}$ with $\sigma_1\geq\cdots\geq\sigma_d\geq0$, each eigenvalue $\sigma_a$ is the probe variance along the orthonormal direction $\mathbf u_a$ \citep{jolliffe2002principal,golub2013matrix}.
The trajectory $\{\Sigma_{k,t}\}_{t}$ therefore constitutes the learning dynamics of geometric representations for the student in MOPD paradigm.
Specifically, its eigenvalues record temporally on how the pushed variance redistributes over training, while its eigenvectors record spatially on where the variation within and across domains lives and overlaps.
Overall, the three metrics below read exactly these two faces of one matrix, the eigenvalues along the temporal dimension and the eigenspaces along the spatial dimension.

\paragraph{Effective dimensionality $\mathrm{eRank}$.}
To measure the breadth of the accumulated influence field, in other words how many feature directions the pushes through the $J$-, $K$-, and $G$-roles of Eq.~\eqref{eq:influence} have spread variance into,
we normalize the spectrum of $\Sigma_{k,t}$ by total variance to  $\rho_a=\sigma_a/\sum_{b=1}^{d}\sigma_b$, and the exponential of the von Neumann entropy of $\rho$ defines the effective rank \citep{roy2007effective, garrido2023rankme, li2025tracing}:
\begin{equation}
\label{eq:erank}
\operatorname{eRank} := \exp \Big({S(\hat{\Sigma}_{k,t})} \Big)=\exp\!\left(-\sum_{a=1}^{d}\rho_a\log\rho_a\right),
\end{equation}
with $0\log0:=0$. $\mathrm{eRank}$ equals $r$ when variance spreads uniformly over $r$ nonzero directions, so it counts the directions the student effectively uses. 
Formally, its rising marks expansion, variance invading new directions, and its falling marks consolidation, variance withdrawing into fewer directions. 
Therefore in MOPD, $\mathrm{eRank}$ traces when each domain expands and consolidates its representation within the student for the temporal signature of the learning progress.

\vspace{-2mm}
\paragraph{Spectral decay $\alpha_{\mathrm{ReQ}}$.}
We trace it to measure the packing of the accumulated influence field, in other words how steeply the variance concentrates from the leading to the trailing directions.
Empirical covariance spectra follow a power law $\sigma_a\propto a^{-\alpha}$ \citep{li2025tracing,agrawal2022areq}, and least squares in log--log coordinates over the numerically meaningful indices \mbox{$I=\{a:\sigma_a>10^{-12}\sigma_1\}$} gives:
\begin{equation}
\label{eq:areq}
\alpha_{\mathrm{ReQ}}=-\frac{\sum_{a\in I}\big(\log a-\bar l\big)\big(\log\sigma_a-\bar s\big)}{\sum_{a\in I}\big(\log a-\bar l\big)^{2}},
\end{equation}
where $\bar l$ and $\bar s$ average $\log a$ and $\log\sigma_a$ over $I$.
Slower decay or smaller $\alpha_{\mathrm{ReQ}}$ implies a more uniform spread of $\sigma_a$ with a higher entropy.
Conversely, faster decay or larger $\alpha_{\mathrm{ReQ}}$ implies the representations are compactly packed along fewer principal directions \citep{agrawal2022areq,stringer2019high}, yielding a lower entropy.

\vspace{-2mm}
\paragraph{Subspace alignment $\operatorname{Sim}_{K_s}$.}
For spatial tracing, we measure the position of the accumulated influence fields, where the pushes settled neglecting by the rotation-invariant spectra above.
Specifically, let $\mathbf Q_{k,t}$ collect the top-$K_s$ eigenvectors of $\Sigma_{k,t}$, with $K_s=16$ throughout, distinct from the number of domains $K$. 
With $\mathbf u_a$ the columns of $\mathbf Q_{k,t}$ and $\mathbf v_b$ those of $\mathbf Q_{k',t'}$, we define: 
\begin{equation}
\label{eq:simk}
\operatorname{Sim}_{K_s}\big((k,t),(k',t')\big)=\frac{1}{K_s}\sum_{a=1}^{K_s}\sum_{b=1}^{K_s}\big(\mathbf u_a^{\top}\mathbf v_b\big)^{2},
\end{equation}
which is the mean squared cosine of the principal angles between the two leading spans \citep{bjorck1973angles, kornblith2019similarity}.
Zero for orthogonal spans and one for coincident spans. 
Its two readings supply the spatial tracing in MOPD.
With $k=k'$ and $t'\neq t$, it measures within-domain persistence on how firmly a domain follows its dominant subspace across training.
Conversely with $k\neq k'$ and $t'=t$, it measures cross-domain overlap on how much of their dominant subspaces the domains share within the student.

\subsection{\textsc{Janus} Perspective to Trace the \textsc{Ianua}} 
Overall, based on the learning dynamics of geometric representations, we propose a formulation and methodology that traces how the student internally learns the multi-domain capabilities.
We claim that the internal performance of the student is the underlying problem under the MOPD paradigm, rather than merely the external teacher supervision and the post-training performance evaluation.
Next, beyond the small-scale and weakly heterogeneous domains of existing MOPD methods such as math, code, and instruction following, we conduct extensive experiments on broader and more practical domains, \emph{e.g.}, Math, Code, Science, ToolUse, and SciSafety, while providing a \textsc{Janus} perspective to trace the \textsc{Ianua} of learning for the model itself.

\section{Temporal Tracing: Learning Dynamics of Multi-Domain Capabilities}
\label{sec:temporal}

In this section, we answer the first question which is on temporal dimension.
Specifically, we first introduce the experimental settings (Section~\ref{sec:temporal settings}). 
We compare MOPD with joint RL to characterize learning progression (Section~\ref{sec:progress}), then combine single-domain controls and domain interventions to examine domain-specific progression and capability trade-offs (Section~\ref{sec:efficiency}).

\begin{figure*}[t]
  \centering
  \centering
  \fontsize{8pt}{9pt}\selectfont
  \setlength{\tabcolsep}{3pt}
  \renewcommand{\arraystretch}{1.12}
  \par\medskip
  \centering
  \vspace{-2mm}
  \includegraphics[width=1.0\textwidth]{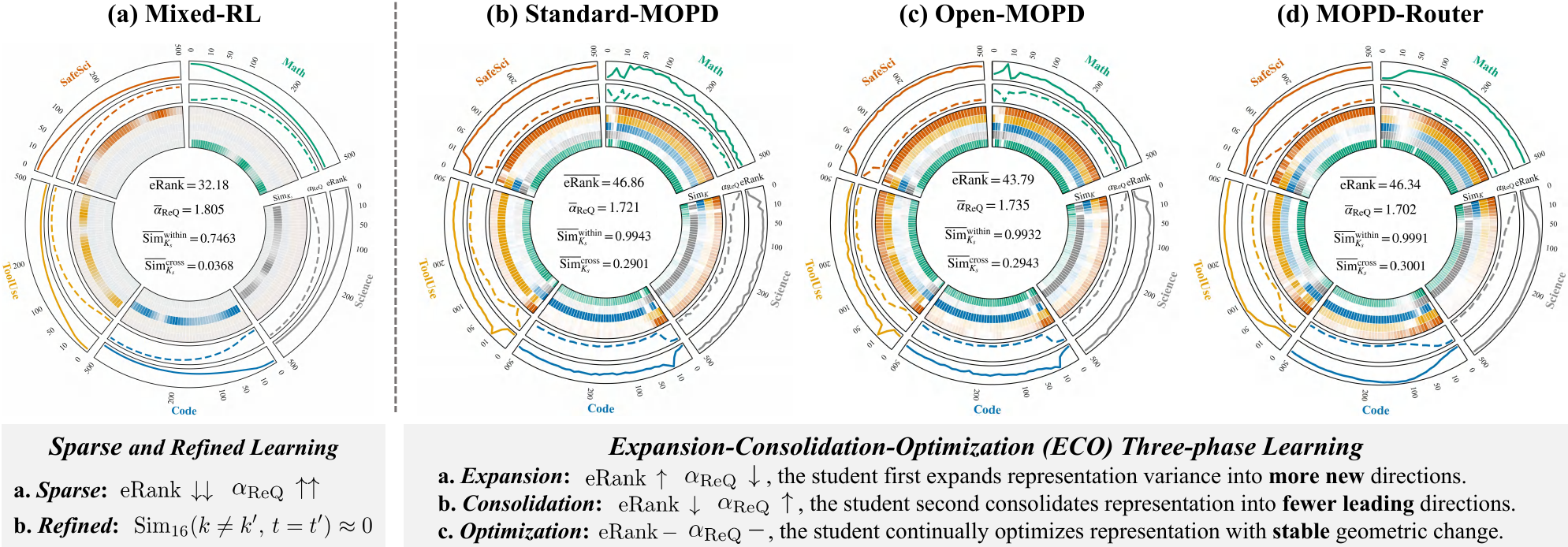}
  \vspace{-5mm}
  \caption{\textbf{Representation geometry across post-training methods.}
  (a) Mixed-RL; (b) Standard-MOPD; (c) Open-MOPD; (d) MOPD-Router.
  Each sector represents one domain: solid and dashed curves show $\mathrm{eRank}$ and $\alpha_{\mathrm{ReQ}}$, respectively, and inner heatmaps show $\operatorname{Sim}_{16}$.
  }
  \label{fig:T.1}
\end{figure*}

\vspace{-2mm}
\subsection{Experimental Setup}
\label{sec:temporal settings}

In this part, we provide the detailed experimental setup, including backbone models, baseline methods, training and evaluation, and implementation details.

\begin{figure*}[!t]
  \centering
  \begingroup
  \fontsize{8pt}{9.5pt}\selectfont
  \setlength{\tabcolsep}{1.5pt}
  \renewcommand{\arraystretch}{1.18}

  \newcommand{\tval}[1]{%
    {\fontsize{7pt}{8.5pt}\selectfont\ensuremath{#1}}%
  }

  \definecolor{tableMin}{HTML}{E7F1FA}
  \definecolor{tableMax}{HTML}{FCE8E6}
  \newcommand{\tmin}[1]{\cellcolor{tableMin}\tval{#1}}
  \newcommand{\tmax}[1]{\cellcolor{tableMax}\tval{#1}}

  \captionof{table}{Validation results (\%) for Qwen3-4B-Instruct and Safin1-4B.
  Light red and light blue mark column maxima and minima among the three MOPD methods, separately for each backbone.}
  \label{tab:qwen3-five-domain-results}
  \vspace{-3mm}

  \begin{tabularx}{\textwidth}{@{}l*{11}{>{\centering\arraybackslash}X}@{}}
    \toprule
    \multirow{2}{*}{Method}
    & \multicolumn{3}{c}{Math}
    & \multicolumn{2}{c}{Science}
    & \multicolumn{2}{c}{Code}
    & ToolUse & SafeSci
    & \multicolumn{2}{c}{Overall} \\[-0.5pt]
    \cmidrule(lr){2-4}
    \cmidrule(lr){5-6}
    \cmidrule(lr){7-8}
    \cmidrule(lr){9-9}
    \cmidrule(lr){10-10}
    \cmidrule(lr){11-12}
    & AIME24 & AIME25 & AIME26
    & SciBench & GPQA
    & LCB-v5 & LCB-v6
    & ToolCall & SciSafety
    & \emph{Avg.} & \emph{Recovery} \\
    \midrule

    \rowcolor{black!6}
    \emph{Qwen3-4B-Instruct (16K)}
    & \tval{57.71} & \tval{45.00} & \tval{50.21}
    & \tval{44.31} & \tval{51.01}
    & \tval{38.00} & \tval{36.05}
    & \tval{46.98} & \tval{50.20}
    & \tval{46.58} & \tval{-} \\

    \hspace{1em}Single-domain GRPO
    & \tval{67.92} & \tval{54.58} & \tval{59.58}
    & \tval{51.17} & \tval{62.42}
    & \tval{58.93} & \tval{57.36}
    & \tval{73.06} & \tval{62.40}
    & \tval{62.21} & \tval{100.00\%} \\

    \hspace{1em}Single-domain OPD
    & \tval{63.33} & \tval{53.54} & \tval{57.50}
    & \tval{51.09} & \tval{62.53}
    & \tval{58.04} & \tval{56.84}
    & \tval{72.18} & \tval{61.00}
    & \tval{61.11} & \tval{92.94\%} \\

    \hspace{1em}Mixed-GRPO
    & \tval{65.42} & \tval{53.33} & \tval{58.13}
    & \tval{50.52} & \tval{61.58}
    & \tval{55.24} & \tval{55.29}
    & \tval{70.97} & \tval{60.00}
    & \tval{60.25} & \tval{87.43\%} \\
    \hdashline

    \hspace{1em}Standard-MOPD
    & \tmin{61.04} & \tmin{50.21} & \tmin{53.96}
    & \tmin{50.17} & \tmax{59.60}
    & \tmax{55.35} & \tmax{53.76}
    & \tval{70.77} & \tval{59.00}
    & \tval{58.86} & \tval{78.52\%} \\

    \hspace{1em}Open-MOPD
    & \tmax{61.67}
    & \tmax{51.46}
    & \tmax{55.42}
    & \tval{50.69} & \tval{57.07}
    & \tval{51.76} & \tval{52.95}
    & \tmax{72.78} & \tmax{60.40}
    & \tmax{59.12} & \tmax{80.21\%} \\

    \hspace{1em}MOPD-Router
    & \tval{61.25} & \tval{50.83} & \tmin{53.96}
    & \tmax{51.21} & \tmin{55.05}
    & \tmin{49.45} & \tmin{52.24}
    & \tmin{57.90} & \tmin{56.80}
    & \tmin{54.82} & \tmin{52.70\%} \\
    \midrule

    \rowcolor{black!6}
    \emph{Safin1-4B (32K)}
    & \tval{69.17} & \tval{63.54} & \tval{55.42}
    & \tval{46.55} & \tval{48.48}
    & \tval{27.00} & \tval{29.90}
    & \tval{55.65} & \tval{52.80}
    & \tval{49.43} & \tval{-} \\

    \hspace{1em}Single-domain GRPO
    & \tval{78.96} & \tval{72.50} & \tval{66.88}
    & \tval{65.00} & \tval{54.55}
    & \tval{43.10} & \tval{43.35}
    & \tval{71.98} & \tval{74.20}
    & \tval{64.39} & \tval{100.00\%} \\

    \hspace{1em}Single-domain OPD
    & \tval{78.75} & \tval{72.08} & \tval{65.63}
    & \tval{64.52} & \tval{52.80}
    & \tval{43.13} & \tval{41.29}
    & \tval{71.98} & \tval{73.60}
    & \tval{63.72} & \tval{95.55\%} \\

    \hspace{1em}Mixed-GRPO
    & \tval{78.75} & \tval{72.50} & \tval{66.67}
    & \tval{65.47} & \tval{53.54}
    & \tval{42.08} & \tval{42.67}
    & \tval{71.45} & \tval{73.20}
    & \tval{63.85} & \tval{96.43\%} \\
    \hdashline

    \hspace{1em}Standard-MOPD
    & \tmax{77.71}
    & \tmax{68.75}
    & \tmax{62.92}
    & \tmin{60.61} & \tmax{53.14}
    & \tmax{41.79} & \tmax{42.00}
    & \tval{70.89} & \tmax{71.80}
    & \tmax{62.25} & \tmax{85.72\%} \\

    \hspace{1em}Open-MOPD
    & \tmin{76.88} & \tval{68.13} & \tval{62.50}
    & \tmax{61.58} & \tval{52.62}
    & \tval{41.08} & \tval{41.44}
    & \tmax{71.91} & \tval{70.80}
    & \tval{62.04} & \tval{84.31\%} \\

    \hspace{1em}MOPD-Router
    & \tval{77.08} & \tmin{67.29} & \tmin{61.25}
    & \tval{60.96} & \tmin{50.20}
    & \tmin{40.35} & \tmin{40.20}
    & \tmin{63.33} & \tmin{68.60}
    & \tmin{59.25} & \tmin{65.68\%} \\
    \bottomrule
  \end{tabularx}

  \par\smallskip
  \begin{minipage}{\textwidth}
    \scriptsize
    Recovery is
    $100(\bar A-\bar A_{\mathrm{backbone}})/
    (\bar A_{\mathrm{SD\text{-}GRPO}}-\bar A_{\mathrm{backbone}})$,
    where each $\bar A$ first averages benchmarks within each domain
    and then averages the five domains.
    The Single-domain GRPO and OPD averages each aggregate five independent
    domain models, rather than one multi-domain model.
    Recovery uses the Single-domain GRPO average for the same backbone.
  \end{minipage}
  \endgroup

  \par\vspace{2mm}
  \includegraphics[
    width=\textwidth,
    height=.38\textheight,
    keepaspectratio
  ]{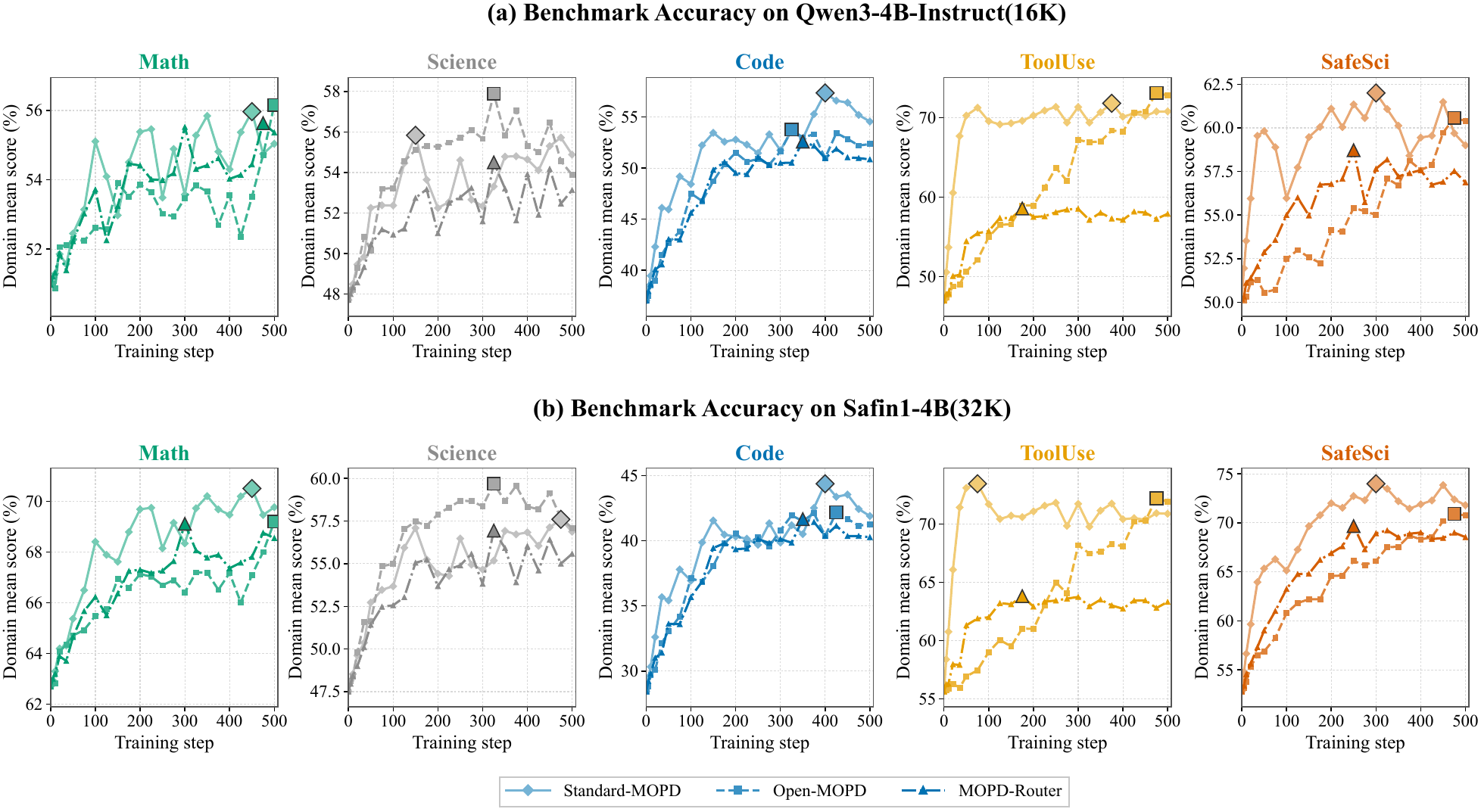}
  \par\vspace{-2mm}
  \caption{\textbf{Benchmark accuracy trajectories across five domains.}
  (a) Qwen3-4B-Instruct (16K); (b) Safin1-4B (32K).
  Each domain score averages its benchmarks,
  following Table~\ref{tab:qwen3-five-domain-results}.}
  \label{fig:T.3}
\end{figure*}

\paragraph{Backbone Models.}

We adopt Qwen3-4B-Instruct-2507~\citep{qwen3_2025} and Safin1-4B~\citep{zhang2026march,zhang2026safin1,zheng2026safin15} as student backbones.
For each backbone, the teacher pool comprises five models obtained through domain-specific RL post-training.
Table~\ref{tab:qwen3-five-domain-results} specifies the response-length limits for each backbone.

\vspace{-3mm}
\paragraph{Baseline Methods.}
We trace three MOPD methods that share the initial student, teacher pool, and training data.
These methods instantiate the temporal coefficient $p_{k,t}$ and spatial coefficient $w_{i,t}(x,y)$ of Eq.~\eqref{eq:mopd-general} as follows. Complete derivations of the mappings are provided in Appendix~\ref{app:experiments-baselines}.
\vspace{-2mm}
\begin{itemize}
 \item \textbf{Standard-MOPD~\citep{mimov2flash2026}}: the vanilla protocol adopted by current frontier pipelines, which uses a fixed domain mixture and hard teacher routing ($p_{k,t}\equiv p_k$, $w_{i,t}=\mathbf 1[i=k]$).
 \vspace{1mm}
 \item \textbf{Open-MOPD~\citep{openmopd2026}}: It coordinates temporal allocation through token-share and gap-dependent loss weighting, yielding adaptive effective domain coefficients $p_{k,t}$.
 \vspace{1mm}
 \item \textbf{MOPD-Router~\citep{mopdrouter2026}}: It coordinates spatial teacher supervision through token-level, direction-dependent weights $w_{i,t}(x,y)$, allowing eligible teachers to jointly supervise each rollout.
\end{itemize}
\vspace{-2mm}

Furthermore, the single-domain OPD controls distill each teacher alone on its own domain and serve as the independent reference for what a domain can reach without cross-domain interference.

\vspace{-2mm}
\paragraph{Training and Evaluation}
We study joint post-training across five domains: mathematical reasoning (Math), code generation (Code), scientific question answering (Science), tool use (ToolUse), and scientific safety (SafeSci).
Each domain contributes 10k prompts to a mixed training set of 50k prompts.
We evaluate Math on AIME24~\citep{h4aime2024}, AIME25~\citep{opencompassaime2025}, and AIME26~\citep{opencompassaime2026}; Science on SciBench~\citep{wang2023scibench} and GPQA~\citep{rein2023gpqa}; Code on LiveCodeBench v5 and v6~\citep{jain2024livecodebench}; ToolUse on a tool-call subset of ToolAlpaca~\citep{tang2023toolalpaca}; and SafeSci on the domain-specific questions of SciSafetyBench~\citep{zhu2025safescientist}.
For Math, we sample 16 responses per problem and report Avg@16 accuracy. The other benchmarks use one sampled response per problem; Code tasks use Pass@1.
Overall \emph{Avg.} first averages benchmarks within each domain and then averages the five domain scores.
Appendix~\ref{app:experiments} lists the complete hyperparameters.

\vspace{-2mm}
\paragraph{Implementation Details.}
The main method comparisons use the same learning rate and shared training hyperparameters.
For each method, we trace 42 non-uniform anchors and compute the observation matrix $\Sigma_{k,t}$ and its three geometric metrics at each anchor.
For each domain, the probes are 10,000 fixed prompts from the training parquet, with measurements averaged over eight sampling repetitions at each anchor.
Geometric trajectories use $\log(1+t/10)$ spacing.
Appendix~\ref{app:tracing-protocol} details the anchor schedule, probe construction, and estimator settings.

\begin{figure*}[!t]
  \centering
  \includegraphics[
    width=\textwidth,
    height=.38\textheight,
    keepaspectratio
  ]{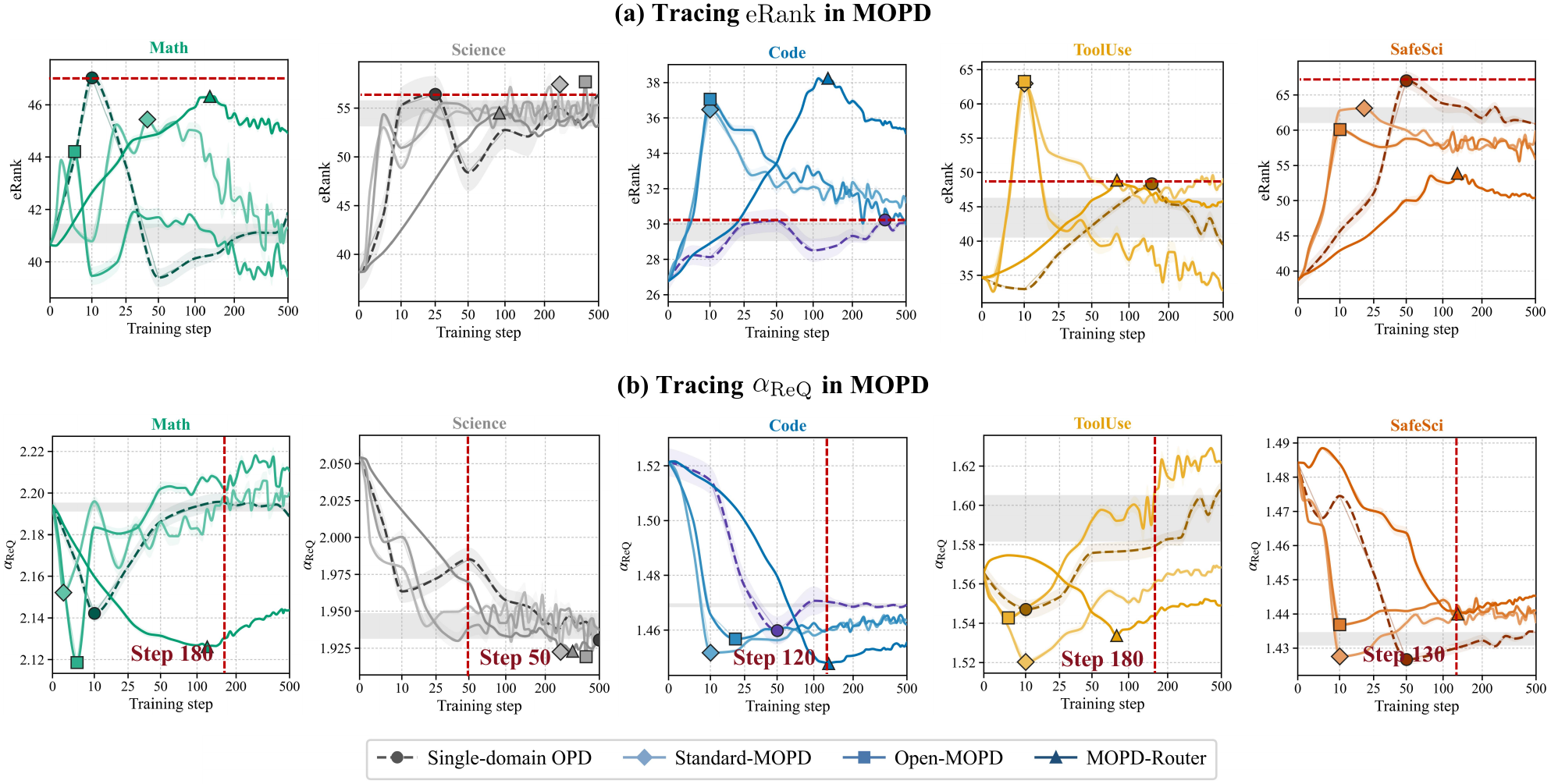}
  \par\vspace{-2mm}
  \captionof{figure}{\textbf{Domain-specific temporal dynamics.}
  (a) $\mathrm{eRank}$; (b) $\alpha_{\mathrm{ReQ}}$.
  Outlined symbols mark $\mathrm{eRank}$ maxima and spectral-decay minima.
  Gray bands show the terminal P10--P90 range of single-domain OPD.
  Red horizontal guides mark its $\mathrm{eRank}$ maximum, red vertical guides mark its stabilization steps.}
  \label{fig:T.2}
\end{figure*}

\subsection{Three-phase Learning Progression within the Student}
\label{sec:progress}

We first characterize the internal learning progression of the student during multi-domain capability acquisition.
Specifically, Figure~\ref{fig:T.1} compares the three MOPD methods with Mixed-GRPO.

Figure~\ref{fig:T.1} jointly reports spectral complexity and subspace alignment from the observation matrices $\Sigma_{k,t}$ across five domains.
Mixed-RL in panel~(a) exhibits decreasing $\mathrm{eRank}$ and increasing $\alpha_{\mathrm{ReQ}}$, consistent with the representation compression reported in prior RLVR analyses \citep{li2025tracing}.
Variance becomes concentrated in fewer effective directions, while low cross-domain $\operatorname{Sim}_{16}$ indicates largely separated dominant subspaces.
Conversely, the three MOPD methods in panels~(b)--(d) exhibit initial representation expansion, subsequent consolidation, and slower late geometric change.
This progression motivates the analogy to ``students'', who first develop a broader representation of the supervised domains and subsequently reorganize and refine it.

We characterize this pattern as an \emph{Expansion--Consolidation--Optimization} (ECO) learning progression.
During \emph{Expansion}, rising $\mathrm{eRank}$ indicates that variance spreads over more effectively used directions, typically accompanied by decreasing $\alpha_{\mathrm{ReQ}}$.
Subsequently, \emph{Consolidation} concentrates variance as $\mathrm{eRank}$ falls from its peak, followed by \emph{Optimization} with slower spectral change and residual drift.
External supervision coordination changes the timing and magnitude of this progression.
Open-MOPD adjusts $p_{k,t}$, with sharper early changes and stronger subsequent contraction in domains such as ToolUse.
MOPD-Router adjusts $w_{i,t}(x,y)$, with more sustained expansion in several domains.
The learning-rate ablation in Appendix~\ref{app:learning-rate} (Figure~\ref{fig:app-lr-temporal}(a)) further shows that the update scale $\eta$ in Eq.~\eqref{eq:update} changes expansion timing, peak $\mathrm{eRank}$, and late variability while retaining the ECO progression.
Figure~\ref{fig:T.1} further show that spectral reorganization coexists with persistent dominant subspaces.

\begin{figure*}[!t]
  \centering
    \centering
  \begingroup
  \definecolor{interventionMin}{HTML}{E7F1FA}
  \definecolor{interventionMax}{HTML}{FCE8E6}
  \fontsize{8pt}{9.5pt}\selectfont
  \setlength{\tabcolsep}{1.5pt}
  \renewcommand{\arraystretch}{1.18}
  \newcommand{\tval}[1]{%
    {\fontsize{7pt}{8.5pt}\selectfont\ensuremath{#1}}%
  }
  \captionof{table}{Validation results (\%) under domain interventions.
  Light blue and light red mark column minima and maxima, respectively.
  Overall is the five-domain macro-average.}
  \label{tab:domain-intervention-results}
  \vspace{-3mm}
  \begin{tabularx}{\textwidth}{@{}l*{10}{>{\centering\arraybackslash}X}@{}}
    \toprule
    \multirow{2}{*}{Method}
    & \multicolumn{3}{c}{Math}
    & \multicolumn{2}{c}{Science}
    & \multicolumn{2}{c}{Code}
    & ToolUse & SafeSci & Overall \\[-0.5pt]
    \cmidrule(lr){2-4}
    \cmidrule(lr){5-6}
    \cmidrule(lr){7-8}
    \cmidrule(lr){9-9}
    \cmidrule(lr){10-10}
    \cmidrule(lr){11-11}
    & AIME24 & AIME25 & AIME26
    & SciBench & GPQA
    & LCB-v5 & LCB-v6
    & ToolCall & SciSafety & \emph{Avg.} \\
    \midrule
    \emph{Qwen3-4B-Instruct (16K)}
    & \tval{57.71} & \tval{45.00} & \tval{50.21}
    & \cellcolor{interventionMin}\tval{44.31} & \cellcolor{interventionMin}\tval{51.11}
    & \tval{38.00} & \tval{36.05}
    & \tval{46.98} & \cellcolor{interventionMin}\tval{50.20}
    & \cellcolor{interventionMin}\tval{46.58} \\
    Standard-MOPD
    & \tval{61.04} & \tval{50.21} & \tval{53.96}
    & \tval{50.17} & \tval{59.60}
    & \tval{55.35} & \tval{53.76}
    & \tval{70.77} & \tval{59.00} & \tval{58.86} \\
    \hdashline
    \hspace{1em}w/o Math
    & \cellcolor{interventionMin}\tval{54.17} & \cellcolor{interventionMin}\tval{43.54} & \cellcolor{interventionMin}\tval{49.79}
    & \tval{50.00} & \tval{60.10}
    & \cellcolor{interventionMax}\tval{56.01} & \cellcolor{interventionMax}\tval{57.33}
    & \tval{71.17} & \tval{60.20} & \tval{58.45} \\
    \hspace{1em}w/o Science
    & \tval{59.58} & \tval{47.50} & \tval{53.75}
    & \cellcolor{interventionMax}\tval{51.90} & \cellcolor{interventionMax}\tval{63.64}
    & \tval{55.20} & \tval{53.24}
    & \cellcolor{interventionMax}\tval{71.77} & \tval{59.20} & \cellcolor{interventionMax}\tval{59.31} \\
    \hspace{1em}w/o Code
    & \cellcolor{interventionMax}\tval{62.08} & \tval{50.83} & \tval{52.29}
    & \tval{49.48} & \tval{58.59}
    & \cellcolor{interventionMin}\tval{37.46} & \cellcolor{interventionMin}\tval{33.48}
    & \tval{69.96} & \cellcolor{interventionMax}\tval{61.20} & \tval{55.15} \\
    \hspace{1em}w/o ToolUse
    & \tval{60.21} & \tval{50.83} & \tval{54.38}
    & \tval{50.17} & \tval{59.95}
    & \tval{54.21} & \tval{54.48}
    & \cellcolor{interventionMin}\tval{29.03} & \tval{60.60} & \tval{50.84} \\
    \hspace{1em}w/o SafeSci
    & \tval{61.88} & \cellcolor{interventionMax}\tval{51.04} & \cellcolor{interventionMax}\tval{54.79}
    & \tval{49.83} & \tval{57.07}
    & \tval{55.70} & \tval{54.57}
    & \tval{71.17} & \tval{53.80} & \tval{57.89} \\
    \bottomrule
  \end{tabularx}
  \endgroup

  \par\vspace{2mm}
  \includegraphics[
    width=\textwidth,
    height=.38\textheight,
    keepaspectratio
  ]{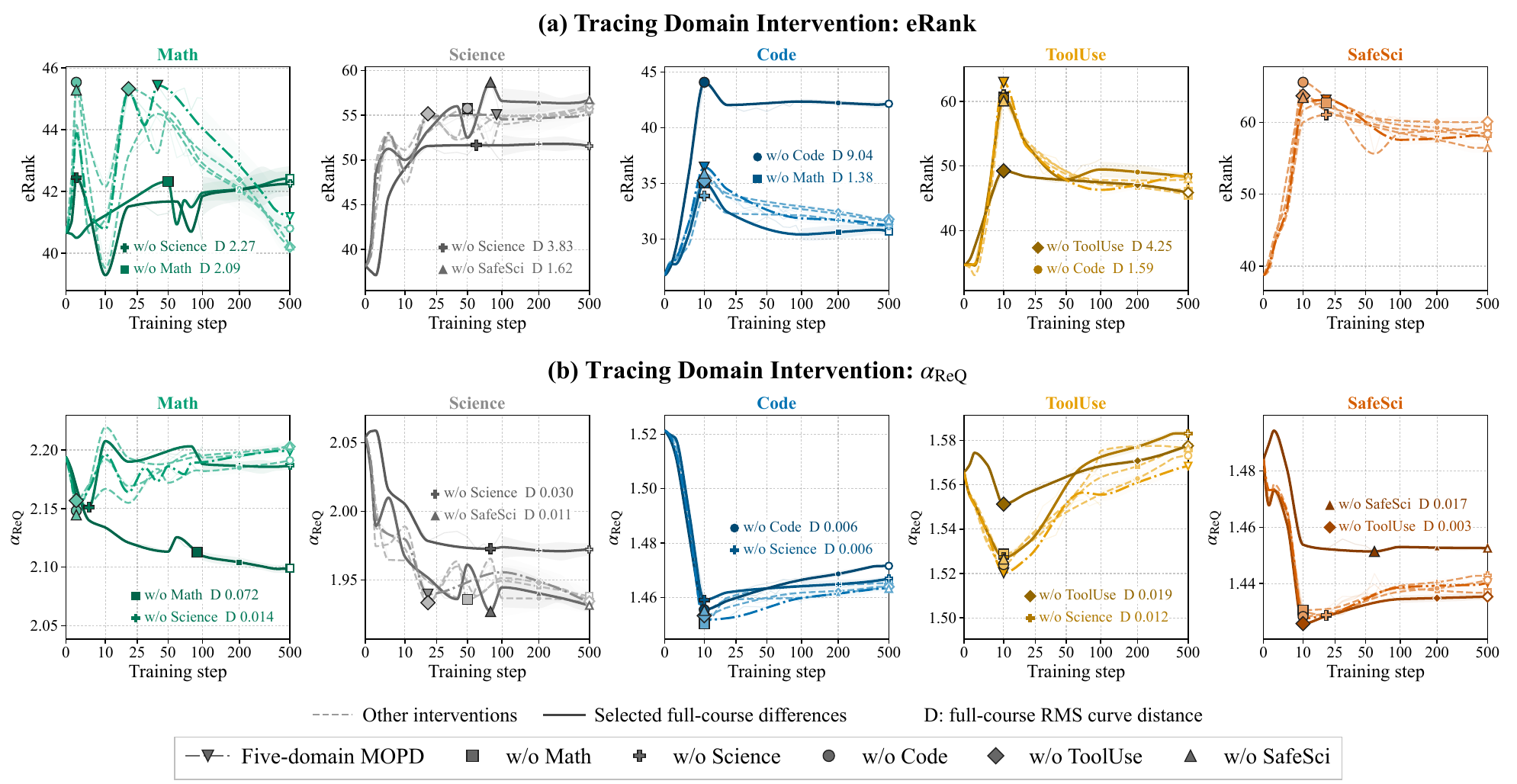}
  \par\vspace{-2mm}
  \captionof{figure}{\textbf{Temporal responses to domain interventions.}
  Columns denote probe domains; rows show (a) $\mathrm{eRank}$ and (b) $\alpha_{\mathrm{ReQ}}$.
  Dark solid curves highlight the largest full-course deviations.}
  \label{fig:T.4}
\end{figure*}

\begin{rqbox}
\textbf{Takeaway 1: The student in MOPD follows an Expansion--Consolidation--Optimization (ECO) learning progression.}
Representation variance first spreads across more effective directions, then concentrates, and finally changes more slowly during continued adaptation.
\end{rqbox}

\subsection{Domain-specific Progression and Capability Trade-offs}
\label{sec:efficiency}

We then study how domain-specific ECO progression relates to capability trade-offs by comparing MOPD with single-domain OPD (Figure~\ref{fig:T.2}).
Specifically, we apply domain-specific interventions that respectively remove one domain and its matched teacher at a fixed total update budget.

As shown in Figure~\ref{fig:T.2}, the method comparisons show domain-specific learning dynamics within a common ECO progression.
Specifically, ToolUse and SafeSci expand earlier in several MOPD runs, whereas Math and Science peak later in others, relative to single-domain OPD.
Math and SafeSci remain below their single-domain $\mathrm{eRank}$ maxima, while Code attains a higher effective dimensionality.
Meanwhile, several late $\alpha_{\mathrm{ReQ}}$ trajectories also remain outside the single-domain terminal ranges, which indicates different consolidation outcomes.
Furthermore, external supervision coordination would redistribute this progression. 
Specifically, Open-MOPD advances the Math $\mathrm{eRank}$ peak, and exhibits stronger subsequent contraction in ToolUse.
MOPD-Router sustains broader Code representations but exhibits later expansion and a smaller ToolUse peak.
The residual capability gaps in Table~\ref{tab:qwen3-five-domain-results} suggest lower scores than single-domain OPD on all three AIME sets for both backbones.
These results show that redistributing the progression does not recover every independently acquired capability.


Domain interventions further suggest that inherent learning limitations in efficiency and scaling cause capabilities across domains to interfere.
As shown in Figure~\ref{fig:T.4}, removing Science weakens the early expansion of math and lowers its accuracy by $1.46$ points, even as Science itself improves by $2.89$ points.
Conversely, removing SafeSci instead broadens the terminal spectrum of Science while lowering its accuracy, revealing a different relation between expansion and benefit.
Code spectrum is comparatively insensitive to other-domain omissions, but its own omission leaves sustained excess expansion and altered $\alpha_{\mathrm{ReQ}}$, with a $19.09$-point accuracy loss that places it below the initial model.
Meanwhile, other-domain updates continue to reshape Code without recovering effective consolidation.
Under inherent learning limitations in efficiency and scaling, removing Science which occupy the larger portion of $\mathrm{eRank}$ achieve the highest overall score.

\begin{rqbox}
\textbf{Takeaway 2: Cross-domain interactions reshape domain-specific progression and create capability trade-offs.}
Shared updates alter the timing and consolidation of domain-specific ECO trajectories, with unequal gains and losses across capabilities.
This coupling limits how effectively independently acquired capabilities can be retained together under a shared training budget.
\end{rqbox}

\section{Spatial Tracing: Organization and Integration of Multi-Domain Capabilities}
\label{sec:spatial}

We study how the student organizes and integrates multi-domain capabilities in representation space.
After describing the experimental setup (Section~\ref{sec:spatial settings}), we trace the dominant subspaces of $\Sigma_{k,t}$ to characterize subspace locking within each domain (Section~\ref{sec:within}) and across domains (Section~\ref{sec:across}).

\subsection{Experimental Setup}
\label{sec:spatial settings}

We use the experimental setup and probes of Section~\ref{sec:temporal settings}, with $K_s=16$ in Eq.~\eqref{eq:simk}.
Within-domain $\operatorname{Sim}_{16}$ compares each current subspace with its own previous saved anchor or initialization, whereas cross-domain  $\operatorname{Sim}_{16}$ averages the overlap of each domain with the other four domains at the same anchor.
These complementary measurements trace the orientation and overlap of the leading eigenspaces of the same observation matrices used for temporal tracing.

\begin{figure*}[t]
  \centering
  \includegraphics[width=\textwidth]{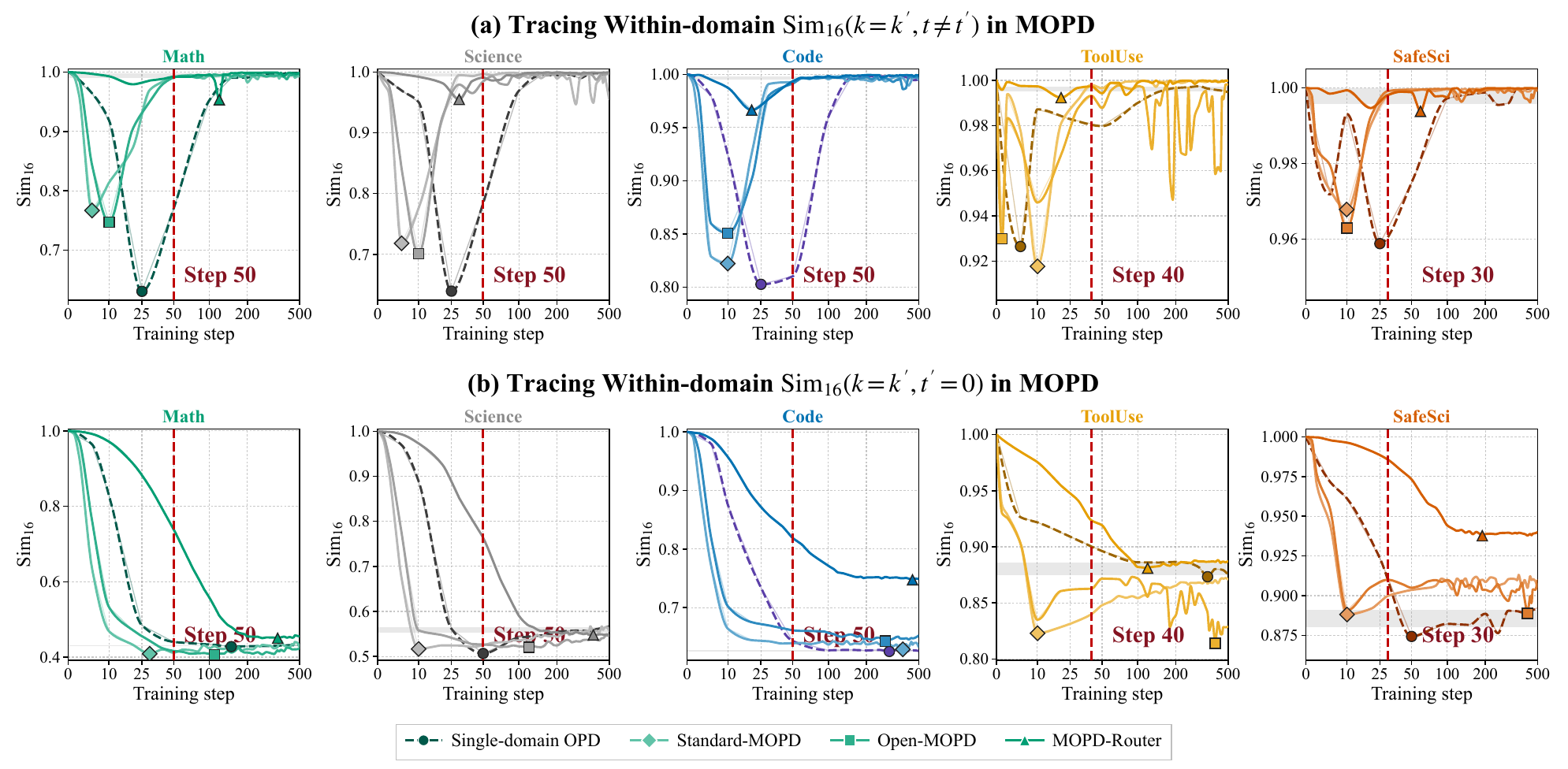}
  \caption{\textbf{Within-domain subspace persistence with two temporal references.}
  (a) The current domain versus its previous saved anchor; (b) the current domain versus its step-0 subspace.}
  \label{fig:S.1}
\end{figure*}

\subsection{Subspace Locking within Each Domain during Learning Progress}
\label{sec:within}

To trace within-domain subspace locking, we compare current dominant subspace with its own previous saved anchor and initialization (Figure~\ref{fig:S.1}).
The first reference measures local persistence, and the second follows cumulative reorientation.
Therefore, We study both the MOPD method comparisons (Figure~\ref{fig:S.3}(a,b)) and the domain intervention in Table~\ref{tab:domain-intervention-results}.

Within-domain trajectories reveal rapid reorganization followed by persistent reuse of the learned dominant span.
Specifically, Math, Science, and Code first move away from initialization as adjacent-anchor similarity drops, then recover high local persistence while retaining this displacement.
ToolUse exhibits more late fluctuation, whereas SafeSci preserves more of its original span.
The two references thus distinguish settling into a new organization from returning to the initial state.
Meanwhile, Open-MOPD and Standard-MOPD reach this regime through sharper early changes, and MOPD-Router reorganizes more gradually.
Across the three methods, mean adjacent-anchor overlap over final steps ranges from $0.9883$ to $0.9994$, while mean overlap with initialization at the final step remains between $0.6718$ and $0.7156$.
Therefore, spectral consolidation coexists with local persistence in a dominant subspace, establishing the within-domain locking pattern.

As shown in Figure~\ref{fig:S.3}, domain interventions show that the spatial interference can barely effect which organization becomes locally persistent.
Specifically, removing Math or Science attenuates its own-initial displacement, yet their capability responses differ: Math loses accuracy while Science improves.
Without Code supervision, the dominant span remains closer to initialization and adjacent-anchor overlap approaches one, even as the spectrum stays broader.
ToolUse provides the complementary case: its representation moves farther from initialization but also becomes locally persistent as accuracy declines.
Therefore, within-domain locking consistently stabilizes a reorganized dominant span undisturbedly, while the spatial interference can barely effect which organization persists and how much specialized capability it retains.

\begin{rqbox}
\textbf{Takeaway 3: Within-domain subspace locking stabilizes a reorganized dominant subspace.}
Early reorientation is followed by persistent reuse of a learned dominant span undisturbedly.
\end{rqbox}

\subsection{Subspace Locking across Domains during Capability Integration}
\label{sec:across}

\begin{figure}[]
  \centering
  \includegraphics[width=\textwidth]{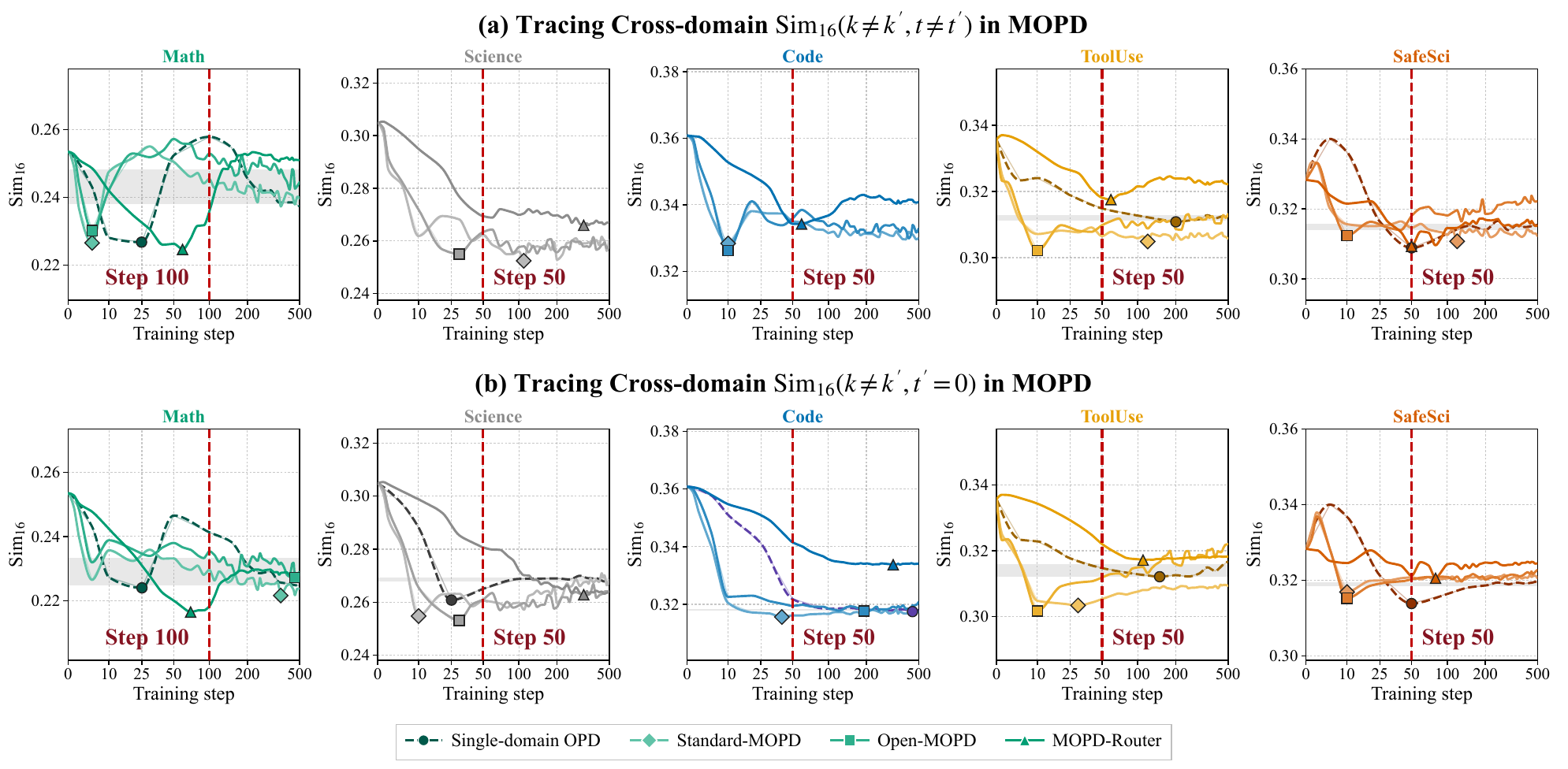}
  \caption{\textbf{Cross-domain subspace overlap with two temporal references.}
  For each domain $k$ at step $t$, (a) averages $\operatorname{Sim}_{16}$ with the other four domains at $t_{\mathrm{prev}}$, and (b) with those domains at step 0.}
  \label{fig:S.2}
\end{figure}

\begin{figure}[!t]
  \centering
  \includegraphics[width=\textwidth]{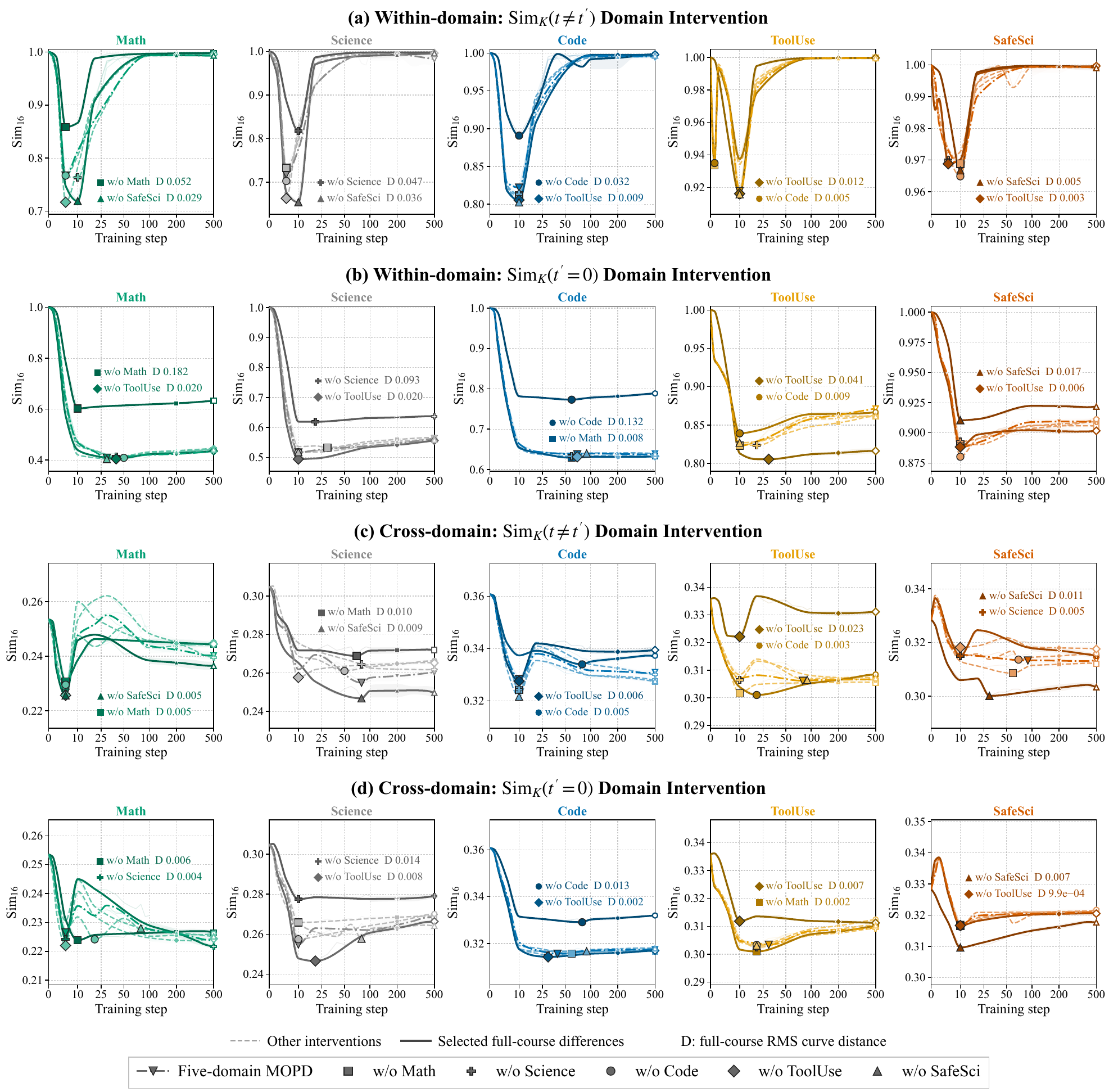}
  \caption{\textbf{Spatial responses to domain interventions.}
  Rows show $\operatorname{Sim}_{16}$ probe previous and initial subspaces within each domain and across domains.}
  \label{fig:S.3}
\end{figure}

To trace cross-domain subspace locking, we compare each current domain with the other four domains' dominant subspaces at the previous anchor and at initialization (Figure~\ref{fig:S.2}).
The previous-anchor reference follows overlap with the evolving multi-domain representation; the initial reference fixes the starting geometry.
We then apply these comparisons to the omission runs (Figure~\ref{fig:S.3}(c,d)) and relate their overlap patterns to the final scores in Table~\ref{tab:domain-intervention-results}.

As shown in Figure~\ref{fig:S.2}, cross-domain trajectories reveal early differentiation followed by persistent, domain-dependent partial overlap.
Specifically, Code retains greater overlap than Math under both references, while Math undergoes a more prolonged, method-dependent adjustment.
Supervision coordination changes the pace of this differentiation: MOPD-Router evolves more gradually, whereas Open-MOPD and Standard-MOPD reorganize more sharply.
Afterward, overlap with both the recent and initial representations of other domains varies more slowly.
Finally, mean initial-reference overlap lies between $0.2879$ and $0.2937$ across the three methods, below the shared initial value of $0.3166$.
Within each method, its range after step 200 is at most $0.0033$.
Therefore, the methods reach stable structures with overlapping and distinct components through different trajectories.

Domain interventions preserve partial overlap while redistributing specialized capability.
As shown in Figure~\ref{fig:S.3}, removing ToolUse raises its overlap with recent subspaces from other domains, barely changes its initial-reference overlap, and reduces accuracy by $41.74$ points.
Code likewise exhibits greater recent-reference overlap and weaker accuracy under its own omission, connecting shared geometric adaptation with incomplete consolidation.
Finally, its overlap rises from $0.330$ to $0.337$, while accuracy falls from $54.56\%$ to $35.47\%$.
In contrast, Science omission increases overlap with other initial spans, while SafeSci omission reduces recent-reference overlap and accuracy together.
Science gains $2.89$ points as initial-reference overlap rises from $0.268$ to $0.280$, whereas SafeSci loses $5.20$ points as recent-reference overlap falls from $0.312$ to $0.306$.
These contrasting responses show that stronger overlap carries unequal learning consequences across domains.
Therefore, cross-domain locking retains a persistent partial-overlap organization across mixtures, and the capability trade-offs depend on which specialization survives within that shared structure.

\begin{rqbox}
\textbf{Takeaway 4: Cross-domain subspace locking retains a stable partial-overlap organization after adaptation.}
Cross-domain comparisons with recent and initial representations settle into
domain-dependent overlap patterns, alongside persistent within-domain subspaces.
\end{rqbox}

\vspace{-2mm}
\section{Discussion and Future Work}
\label{sec:discussion}

\paragraph{Geometric evidence and continual capability integration.}
Janus-MOPD shows distinct temporal and spatial organization under MOPD and Mixed-RL.
After consolidation, aggregate $\mathrm{eRank}$ remains above initialization under MOPD but falls below it under Mixed-RL (Figure~\ref{fig:T.1}), while $\alpha_{\mathrm{ReQ}}$ and $\operatorname{Sim}_{16}$ indicate broader variance and partial overlap among MOPD domain subspaces.
These ECO transitions and subspace-locking patterns suggest a continual learning view.
Specifically, expansion may support plasticity, consolidation and within-domain persistence may support retention, and cross-domain overlap may enable reuse while constraining adaptation to conflicting new domains.
This view connects external supervision allocation through $p_{k,t}$ and $w_{i,t}(x,y)$ to the internal balance of plasticity and retention, but do not directly establish forgetting, causal transfer, or unused capacity~\citep{ren2026continual}.
Testing these interpretations requires sequential domain-arrival and retention evaluations.

\vspace{-2mm}
\paragraph{Internal capability organization beyond external routing.}
Safin-1 provides a state-based example that learned capabilities may be organized and reused through persistent internal states~\citep{zhang2026safin1}.
For MOPD, the question then extends beyond which teacher supplies supervision to how knowledge acquired from several domains remains available, reusable, and sufficiently distinct when tasks require specialization.
This reframes capability organization as part of the learning problem itself.
Specifically, integration must balance reuse with specialization within the student, alongside coordination of external supervision.
The Janus temporal and spatial perspectives can help assess whether such organization evolves over training while preserving shared and domain-specific structure.

\vspace{-2mm}
\paragraph{Agentic MOPD through the Janus Perspective.}
Agentic MOPD must address closed-loop, long-horizon tasks such as tool use and software engineering (SWE), where actions alter later states and terminal outcomes provide sparse credit.
ActionOPD aligns semantic actions at student-visited states, while ToolOPD finds nonuniform, outcome-dependent credit across tool-use trajectories~\citep{actionopd2026,toolopd2026}.
From the Janus perspective, the open question is how planning, tool-use, and feedback representations evolve across interaction stages, and whether their geometry predicts success, recovery, and retention.

\section{Related Work}
\label{sec:related}

\paragraph{Multi-teacher on-policy distillation.}
Integrating multi-domain capabilities into a single model is a fundamental challenge in post-training for large reasoning models (LRMs) \citep{deepseekr1, mimov2flash2026, deepseekv42026}. 
Therefore, multi-teacher on-policy distillation (MOPD)  \citep{mopd2026} was proposed
which integrates the capabilities of multiple domain-specific teacher models into a single student model. 
Specifically, MOPD first trains several teachers separately with domain-specific data, and then applies a domain router to match the responses of student to the corresponding teacher for on-policy
distillation (OPD) \citep{agarwal2024onpolicy, gu2024minillm, lu2025onpolicy}. 
However, the student usually fails to acquire multi-domain capabilities simultaneously due to the undesirable domain heterogeneity \citep{openmopd2026, roadmopd2026, immopd2026}. 
Existing works attribute this phenomenon to the uneven efficiency or hard allocation of the teacher supervision, and propose to coordinate the teacher-student gaps across domains \citep{openmopd2026, roadmopd2026, immopd2026, deltamopd2026} and design soft router for multi-teacher joint supervision \citep{roadmopd2026, mopdrouter2026, trustmopd2026, camopd2026, dnmopd2026} respectively. 
In most cases, we observe that these methods focus on addressing the temporal imbalance and spatial one-sidedness for external teacher supervision, while neglecting the organization and integration of multi-domain capabilities for student itself. 
Conversely, we focus our perspective on the internal performance and evolution of the student, and systematically study how the student learns multi-domain capabilities during MOPD training process.

\paragraph{Learning dynamics of geometric representations.}
Learning dynamics analysis relates parameter updates to model behavior, especially from exact dynamics of linearized networks \citep{saxe2014exact, jacot2018ntk} to the influence decomposition of LLM finetuning \citep{ren2025learning, ren2026continual}.
Meanwhile, the stability--plasticity dilemma in continual learning has long been widely studied through learning dynamics\citep{ren2026continual,french1999catastrophic, kirkpatrick2017overcoming, lyle2023plasticity, dohare2024loss}. 
Technically, mainstream works quantify the geometric evolution of representations with spectral analysis \citep{jolliffe2002principal, golub2013matrix, bjorck1973angles}, through effective rank \citep{roy2007effective, garrido2023rankme, cui2025entropy}, eigenspectrum decay \citep{agrawal2022areq}, and subspace similarity \citep{kornblith2019similarity}. 
Recent work examines post-training shifts in representation geometry \citep{li2025tracing}, OPD generalization \citep{li2026dualnature}, token importance \citep{xu2026tip}, and self-distillation limits \citep{wang2026denser}.
In particular, \citet{shen2026geometryopd} identify early subspace locking of parameter updates in single-domain OPD.
Such isolated-domain results cannot explain temporal interference among jointly learned domains, spatial interactions among their representation subspaces, or how domain allocation and teacher weighting reshape these dynamics.
We therefore study MOPD through a common spectral framework, relating supervision weighting to capability trade-offs and representation organization within the shared student.
\section{Conclusion}
\label{sec:conclusion}

We introduced \textsc{Janus-MOPD}, a unified learning dynamics formulation and geometric tracing methodology for understanding how a student learns and integrates multi-domain capabilities under MOPD.
Tracing $\mathrm{eRank}$, $\alpha_{\mathrm{ReQ}}$, and $\operatorname{Sim}_{K}$ on a common representation covariance matrix connects temporal changes in variance allocation with spatial changes in dominant subspaces.
Across five domains, the three measured MOPD methods exhibit a common ECO progression, heterogeneous domain trajectories, and residual capability trade-offs.
Their representations also develop high local persistence within domains and sustained partial overlap across domains.
Temporal and spatial supervision coordination reshape these dynamics while leaving gaps relative to independent single-domain OPD.
Together, our formulation, methodology, and findings establish the student's internal organization as a central axis for studying MOPD.
This \textsc{Janus} perspective motivates coordinating external supervision with internal representation and adaptation for continual capability integration.

\phantomsection
\addcontentsline{toc}{section}{References}
\begingroup
\sloppy
\bibliography{main}
\endgroup

\addtocontents{toc}{\protect\setcounter{tocdepth}{-1}}
\appendix
\clearpage
\thispagestyle{fancy}

\begingroup
\fontsize{11}{14}\selectfont
\newcommand{\appcontentssection}[2]{%
  \par\addvspace{0.65\baselineskip}\noindent
  \hyperref[#1]{\textbf{\ref*{#1}\quad #2}}\hfill\pageref*{#1}\par}
\newcommand{\appcontentssubsection}[2]{%
  \par\noindent\hspace*{2em}%
  \hyperref[#1]{\ref*{#1}\quad #2}\nobreak\dotfill\nobreak\pageref*{#1}\par}
\phantomsection
\pdfbookmark[1]{Contents of the Appendices}{appendix-contents}
\noindent{\fontsize{18}{22}\selectfont\bfseries Contents of the Appendices}\par
\vspace{1.0\baselineskip}
\appcontentssection{app:theory}{Mathematical Derivations}
\appcontentssubsection{app:theory-visitation}{Reverse-KL Signal and Visitation Convention}
\appcontentssubsection{app:theory-response}{Prediction and Representation Responses}
\appcontentssubsection{app:theory-covariance}{Evolution of Probe Covariance}
\appcontentssection{app:experiments}{More Details of Experiments}
\appcontentssubsection{app:experiments-arms}{Backbone Models and Teacher Pool}
\appcontentssubsection{app:experiments-data}{Tasks and Datasets}
\appcontentssubsection{app:experiments-baselines}{Baseline Methods}
\appcontentssubsection{app:experiments-config}{Hyperparameters and Training Configuration}
\appcontentssubsection{app:experiments-prompts}{Prompt Construction and Reward Functions}
\appcontentssubsection{app:experiments-evaluation}{Evaluation}
\appcontentssubsection{app:experiments-anchors}{Checkpoint and Probe Protocol}
\appcontentssubsection{app:domain-interventions}{Domain-Intervention Protocol and Final Evaluation}
\appcontentssubsection{app:learning-rate}{Learning-Rate Sensitivity}
\appcontentssubsection{app:experiments-coordination}{Supervision Coordination across Five Domains}
\appcontentssection{app:tracing-protocol}{More Algorithmic Analyses}
\appcontentssubsection{app:estimation}{Spectral Estimation and Effective Dimension}
\appcontentssubsection{app:subspaces}{Principal Subspaces and Their Comparison}
\appcontentssubsection{app:reference-geometry}{Previous- and Initial-Reference Geometry}
\appcontentssubsection{app:geometry-robustness}{Numerical Diagnostics and Subspace-Dimension Sensitivity}
\endgroup

\clearpage
\section{Mathematical Derivations}
\label{app:theory}

This appendix makes the local-update convention behind Eq.~\eqref{eq:influence} explicit and connects it to the covariance observed by the geometric probes.

\subsection{Reverse-KL Signal and Visitation Convention}
\label{app:theory-visitation}

At one visited context $s=(x,y_{<j})$, write $q_v=\pi_{\theta_t}(v\mid s)$ and $\tau_v=\pi_{T_k}(v\mid s)$. For the full-vocabulary reverse KL $\ell_k(s)=\sum_v q_v\log(q_v/\tau_v)$, the softmax derivative gives
\begin{equation}
\label{eq:app-rkl-gradient}
\frac{\partial\ell_k}{\partial z_v}
=q_v\left[\log\frac{q_v}{\tau_v}-\ell_k(s)\right].
\end{equation}
The term $\ell_k(s)$ centers the derivative: summing Eq.~\eqref{eq:app-rkl-gradient} over $v$ gives zero. If $\hat A(y)=\log\tau_y-\log q_y$, then
\begin{equation}
\label{eq:app-score-identity}
\mathbb E_{y\sim q}\!\left[\hat A(y)\nabla_{\mathbf z}\log q_y\right]
=-\nabla_{\mathbf z}\ell_k(s).
\end{equation}
Thus the expected sampled-token score update in Eq.~\eqref{eq:update} agrees with descent on the full reverse KL at a fixed visited context. Equation~\eqref{eq:app-rkl-gradient} specifies the exact logit-level signal used in the derivations below.

For the local step we hold the sampled contexts and domain allocation fixed while differentiating the loss evaluated on them. This is the convention used by an optimizer acting on a collected rollout batch. Differentiating the sampling distribution itself would add score-function terms. Let $\mathbf g^{(b)}_{k,t}(s_j)$ be the derivative of the \emph{implemented} token loss with respect to the student logits, including its prescribed stop-gradients and reduction. With $\mathbf J_{z,t}(s_j)=\nabla_\theta\mathbf z_t(s_j)$, a gradient-descent step is
\begin{equation}
\label{eq:app-mopd-update}
\delta\theta_t=-\eta_t\sum_k p_{k,t}\,
\mathbb E_{k,t}\!\left[\sum_j\mathbf J_{z,t}(s_j)^\top
\mathbf g^{(b)}_{k,t}(s_j)\right].
\end{equation}
For detached teacher weights applied to full-distribution losses, $\mathbf g^{(b)}_{k,t}=\sum_iw_{i,t}\nabla_{\mathbf z}\ell_i$. Differentiated weights or clipped sampled-token losses have different derivatives but retain Eq.~\eqref{eq:app-mopd-update}'s chain-rule form. For a stateful optimizer, the executed $\delta\theta_t$ also includes optimizer state and preconditioning; the plain-gradient expression describes the local objective contribution before those transformations.

\subsection{Prediction and Representation Responses}
\label{app:theory-response}

For any differentiable vector observable $\mathbf f_\theta(s_o)$ at a fixed observing context $s_o$, Taylor expansion around $\theta_t$ gives
\begin{equation}
\label{eq:app-observable-response}
\Delta\mathbf f_t(s_o)=\mathbf J_{f,t}(s_o)\delta\theta_t
+O(\|\delta\theta_t\|^2).
\end{equation}
If its coordinate Hessians have bounded operator norm along the step, the remainder in each coordinate is bounded by a constant times $\|\delta\theta_t\|^2$. Inserting Eq.~\eqref{eq:app-mopd-update} yields the source-to-observer kernel $\mathcal K_{f,t}(s_o,s_j)=\mathbf J_{f,t}(s_o)\mathbf J_{z,t}(s_j)^\top$:
\begin{equation}
\label{eq:app-kernel-response}
\Delta\mathbf f_t(s_o)=-\eta_t\sum_kp_{k,t}\,
\mathbb E_{k,t}\!\left[\sum_j
\mathcal K_{f,t}(s_o,s_j)\mathbf g^{(b)}_{k,t}(s_j)\right]
+O(\|\delta\theta_t\|^2).
\end{equation}
For $\mathbf f=\log\boldsymbol\pi_\theta$, the observing Jacobian is the log-softmax Jacobian times $\mathbf J_z$, giving the log-probability response corresponding to Eq.~\eqref{eq:influence} after selecting a token coordinate. For $\mathbf f=\mathbf h_\theta$, the same update is read through the hidden-state Jacobian instead. These are two observables of one parameter step.

\subsection{Evolution of Probe Covariance}
\label{app:theory-covariance}

Stack the $M$ representations of domain $k$ as rows of $\mathbf F_{k,t}$, and define the centering matrix $\mathbf C_M=\mathbf I_M-M^{-1}\mathbf1\mathbf1^\top$ and $\mathbf H_{k,t}=\mathbf C_M\mathbf F_{k,t}$. Equation~\eqref{eq:feature-covariance} then reads $\Sigma_{k,t}=M^{-1}\mathbf H_{k,t}^\top\mathbf H_{k,t}$. Writing $\Delta\mathbf H_{k,t}=\mathbf H_{k,t+1}-\mathbf H_{k,t}$ gives the exact identity
\begin{equation}
\label{eq:app-covariance-change}
\Delta\Sigma_{k,t}=\frac{1}{M}\left(
\mathbf H_{k,t}^\top\Delta\mathbf H_{k,t}
+\Delta\mathbf H_{k,t}^\top\mathbf H_{k,t}
+\Delta\mathbf H_{k,t}^\top\Delta\mathbf H_{k,t}\right).
\end{equation}
Each row of $\Delta\mathbf H$ is a centered representation response of Eq.~\eqref{eq:app-observable-response}. Consequently, covariance changes collect the update influence over a fixed probe population. Eigenvalues describe the redistribution of variance, while eigenvectors locate its dominant directions.

\section{More Details of Experiments}
\label{app:experiments}

\subsection{Backbone Models and Teacher Pool}
\label{app:experiments-arms}

We study the acquisition of five-domain capabilities by a shared student, using Qwen3-4B-Instruct-2507~\citep{qwen3_2025} for the main tracing comparison.
This backbone has 36 transformer blocks, a hidden dimension of 2,560, and a shared tokenizer across the student and its domain experts.
Five independently RL-trained experts provide Math, Science, Code, ToolUse, and SafeSci supervision.
The experts remain frozen during distillation, and the three MOPD methods share the same student initialization and teacher pool.
Table~\ref{tab:app-models} summarizes their roles.
Safin1-4B~\citep{zhang2026safin1} provides a complementary architecture with persistent capability states; its benchmark results appear in Table~\ref{tab:qwen3-five-domain-results} and Figure~\ref{fig:T.3}(b).

\begin{table}[htbp]
\centering
\small
\setlength{\tabcolsep}{5pt}
\caption{Models and their roles. Response limits describe capability evaluation; the shared Qwen3 training configuration appears in Table~\ref{tab:app-training-config}.}
\label{tab:app-models}
\begin{tabular}{@{}p{0.20\textwidth}p{0.43\textwidth}p{0.28\textwidth}@{}}
\toprule
\textbf{Role} & \textbf{Model} & \textbf{Setting} \\
\midrule
Main student & Qwen3-4B-Instruct-2507 & 16K response limit \\
Teacher pool & Five RL experts from the same Qwen3 backbone & One expert per domain \\
Single-domain & Five independently distilled Qwen3 students & One domain per student \\
Additional backbone & Safin1-4B & 32K response limit \\
\bottomrule
\end{tabular}
\end{table}

\subsection{Tasks and Datasets}
\label{app:experiments-data}

We jointly train on mathematical reasoning (\textbf{Math}), scientific question answering (\textbf{Science}), code generation (\textbf{Code}), tool use (\textbf{ToolUse}), and scientific safety (\textbf{SafeSci}).
Each record contains a domain identifier, a role-tagged prompt, and the task metadata required by its verifier.
The frozen mixture contains 10,000 training records per domain, totaling 50,000, and is shuffled with seed 42.
Table~\ref{tab:app-datasets} lists the training sources and the nine evaluation benchmarks.

\paragraph{Math.}
The training pool contains 10,000 verifiable problems derived from NuminaMath-1.5~\citep{numina2024dataset}, spanning competition mathematics and quantitative reasoning.
We preserve the problem statements and final-answer fields and evaluate on AIME24, AIME25, and AIME26 ~\citep{h4aime2024,opencompassaime2025,opencompassaime2026}.

\paragraph{Science.}
The training set contains 10,000 multiple-choice questions from Nemotron-RL-knowledge-mcqa~\citep{nvidia2025nemotronmcqa}.
Evaluation uses 580 quantitative SciBench problems~\citep{wang2023scibench} and 198 GPQA-Diamond questions~\citep{rein2023gpqa}, covering both numerical scientific reasoning and multiple-choice reasoning.

\paragraph{Code.}
The frozen programming pool combines 8,460 TACO-derived problems~\citep{li2023taco} with 1,540 LeetCode tasks.
Each problem retains executable tests and its input/output or functional specification.
We evaluate on LiveCodeBench v5 and v6~\citep{jain2024livecodebench}, containing 167 and 175 problems in the saved evaluation splits.

\paragraph{ToolUse.}
ToolACE~\citep{liu2024toolace} supplies function specifications, user requests, and structured call targets.
The 10,000-record training pool contains 9,513 distinct call examples and 487 repeated records retained to balance the domain sizes.
The ToolCall benchmark uses 496 held-out call targets.

\paragraph{SafeSci.}
SafeSciTrain supplies 10,000 scientific-safety questions, and 500 held-out SafeSciBench questions form the SciSafety evaluation~\citep{zhu2025safescientist}.
The tasks require selecting a safe answer under scientific constraints, retaining the domain-specific choices and answer tags.

\begin{table}[htbp]
\centering
\footnotesize
\setlength{\tabcolsep}{5pt}
\renewcommand{\arraystretch}{1.15}
\caption{Training and evaluation data across five domains.}
\label{tab:app-datasets}
\begin{tabular}{@{}lp{0.35\textwidth}p{0.44\textwidth}@{}}
\toprule
\textbf{Domain} & \textbf{Training} & \textbf{Evaluation} \\
\midrule
Math & NuminaMath-1.5 (10,000) & AIME24 / AIME25 / AIME26 (30 each) \\
Science & Nemotron-RL-knowledge-mcqa (10,000) & SciBench (580); GPQA-Diamond (198) \\
Code & TACO-derived (8,460); LeetCode (1,540) & LiveCodeBench v5 (167) / v6 (175) \\
ToolUse & ToolACE (10,000) & ToolCall (496) \\
SafeSci & SafeSciTrain (10,000) & SafeSciBench / SciSafety (500) \\
\bottomrule
\end{tabular}
\end{table}

\subsection{Baseline Methods}
\label{app:experiments-baselines}

The main tracing comparison uses Standard-MOPD, Open-MOPD, and MOPD-Router with the shared backbone, teacher pool, and training mixture.
Single-domain OPD isolates acquisition from one teacher, while Mixed-GRPO provides a joint RL comparison.
D$^3$-MOPD and Trust-MOPD further characterize scheduling and reliability mechanisms under the coefficient system of Eq.~\eqref{eq:mopd-general}.
For the mappings below, $s_j=(x,y_{<j})$ denotes a visited prefix, $w_{i,t}(s_j)$ is the token-level reading of $w_{i,t}(x,y)$, and $\pi_0$ is the common pre-RL policy.
Loss reweighting induces a normalized effective domain coefficient even when the prompt sampler retains $p_k$.
The allocation coefficients are fixed within each policy update; the estimators, loss reductions, and batch normalization remain those of the methods.

\paragraph{Single-domain GRPO and Mixed-GRPO.}
GRPO~\citep{shao2024deepseekmath} optimizes verifier rewards using advantages relative to the response group of each prompt.
Single-domain GRPO yields independently specialized teachers, whereas Mixed-GRPO optimizes one student on the pooled domains.
The single-domain teacher row in Table~\ref{tab:qwen3-five-domain-results} combines independent models evaluated on their respective tasks.

\paragraph{Single-domain OPD.}
Each student generates its own responses and receives dense supervision exclusively from the corresponding frozen domain expert.
The resulting independent controls quantify the capability attainable without joint teacher supervision.
They also supply the terminal spectral ranges used in Figures~\ref{fig:T.2} and~\ref{fig:S.1}.

\paragraph{Standard-MOPD.}
Standard-MOPD~\citep{mimov2flash2026,mopd2026} pools and shuffles the five datasets and assigns each rollout to its matched expert.
Equal dataset sizes give expected prompt shares $p_k=1/5$, while token contributions vary with response length.
This instantiates fixed $p_{k,t}$ and hard teacher assignment $w_{i,t}=\mathbf{1}[i=k]$ in Eq.~\eqref{eq:mopd-general}.

\paragraph{Open-MOPD.}
Open-MOPD~\citep[Sections~4.1--4.3]{openmopd2026} combines token-share balancing with gap-following loss allocation.
For a batch containing all domains, let $\nu_{k,t}>0$ be the observed token share, $p_k^\star$ the target budget with $\sum_kp_k^\star=1$, and $g_{k,t}$ the mean absolute distillation signal.
With $\bar g_t=K^{-1}\sum_dg_{d,t}$, define $b_{k,t}=\operatorname{clip}((g_{k,t}/\bar g_t)^\alpha,0.05,20)$ for $\alpha>0$.
The loss multiplier is proportional to $p_k^\star b_{k,t}/\nu_{k,t}$; grouping token-sum losses by prompt domain and normalizing gives
\begin{equation}
\label{eq:app-open-pw}
\bigl(p_{k,t}^{\mathrm{Open}},w_{i,t}^{\mathrm{Open}}(s_j)\bigr)
=\left(
\frac{p_k p_k^\star b_{k,t}/\nu_{k,t}}
{\sum_{d=1}^{K}p_d p_d^\star b_{d,t}/\nu_{d,t}},\;
\mathbf 1[i=k]
\right).
\end{equation}
Thus, $p_{k,t}^{\mathrm{Open}}$ represents effective domain-loss allocation, while prompt sampling and hard teacher assignment remain fixed.
Reward refresh changes the student-dependent signal inside each domain loss while preserving this allocation rule.
The anchored experimental variant replaces $\bar g_t$ with $g_{k,\mathrm{init}}$, as specified in Eq.~\eqref{eq:app-open-weight}.

\paragraph{MOPD-Router.}
The ExpertAlign variant of MOPD-Router~\citep[Section~3.3]{mopdrouter2026} selects teachers by directional agreement.
At $s_j$, let $\mathbf e_i=\log\pi_{T_i}-\log\pi_0$ and $\mathbf d_{i,t}=\log\pi_{T_i}-\log\pi_{\theta_t}$ be log-probability vectors restricted to the top-16 candidates of the rollout student.
Define $c_{i,t}=[\cos(\mathbf e_i,\mathbf d_{i,t})]_+$, with $[a]_+=\max(a,0)$, and retain $\mathcal E_t=\{i:\langle\mathbf e_i,\mathbf d_{i,t}\rangle>\delta\}$.
Normalizing the eligible cosine scores gives
\begin{equation}
\label{eq:app-router-pw}
\bigl(p_{k,t}^{\mathrm{Router}},w_{i,t}^{\mathrm{Router}}(s_j)\bigr)
=\left(
p_k,\;
\frac{\mathbf 1[i\in\mathcal E_t]c_{i,t}}
{\sum_{r\in\mathcal E_t}c_{r,t}}
\right).
\end{equation}
Zero-norm directions receive zero scores; an empty eligible set gives zero teacher weights and skips supervision at that token.
With $\delta=10^{-6}$, this changes prefix-dependent $w_{i,t}$ while preserving the domain mixture $p_k$.

\needspace{15\baselineskip}
\paragraph{D$^3$-MOPD.}
D$^3$-MOPD~\citep[Section~3]{d3mopd2026} changes the prompt mixture using both remaining gap and recent progress.
Let $L_{k,t}$ be the smoothed reverse-KL and $L_{k,\mathrm{init}}$ its initial observation mean.
Its score $h_{k,t}=(L_{k,t}/L_{k,\mathrm{init}})v_{k,t}$ uses the positive part of the mean relative KL decrease over available nonoverlapping windows, with each decrease normalized by its earlier endpoint.
Max-normalized scores $\widehat h_{k,t}=h_{k,t}/\max_dh_{d,t}$ define a scheduled mixture $q_{k,t}=\epsilon+(1-K\epsilon)\operatorname{softmax}(\widehat{\mathbf h}_t/\tau)_k$, where $0<\epsilon<1/K$ and $\tau>0$.
With batch jitter $u_{k,t}\sim\operatorname{Uniform}(-\zeta,\zeta)$, $0\leq\zeta<1$, the continuous sampling coefficients before quota rounding are
\begin{equation}
\label{eq:app-d3-pw}
\bigl(p_{k,t}^{\mathrm{D^3}},w_{i,t}^{\mathrm{D^3}}(s_j)\bigr)
=\left(
\frac{q_{k,t}(1+u_{k,t})}
{\sum_{d=1}^{K}q_{d,t}(1+u_{d,t})},\;
\mathbf 1[i=k]
\right).
\end{equation}
The target is uniform during warmup or when all scores vanish.
Thus, temporal allocation changes the visited domain mixture, while teacher assignment remains hard.

\paragraph{Trust-MOPD.}
Trust-MOPD~\citep[Sections~3.1--3.3]{trustmopd2026} estimates local reliability from the magnitude of acquired specialization.
At $s_j$, let $\rho_{i,t}$ be the Euclidean norm of $\log\pi_{T_i}-\log\pi_0$ after subtracting its coordinate mean, and let $\mu_i>0$ be its fixed token-weighted mean on calibration rollouts from the initial student.
Power-normalizing the calibrated magnitudes gives, for $\gamma>0$ and nonzero total score,
\begin{equation}
\label{eq:app-trust-pw}
\bigl(p_{k,t}^{\mathrm{Trust}},w_{i,t}^{\mathrm{Trust}}(s_j)\bigr)
=\left(
p_k,\;
\frac{(\rho_{i,t}/\mu_i)^\gamma}
{\sum_{r=1}^{K}(\rho_{r,t}/\mu_r)^\gamma}
\right).
\end{equation}
Calibration removes persistent scale differences among teachers, while $\gamma$ controls allocation concentration.
The method therefore changes label-free, prefix-dependent $w_{i,t}$ with unit total teacher weight, leaving $p_k$ fixed.

\subsection{Hyperparameters and Training Configuration}
\label{app:experiments-config}

The Qwen3 MOPD runs use VERL~\citep{sheng2025hybridflow} on eight NVIDIA H200 GPUs with vLLM rollout generation and FSDP training.
Table~\ref{tab:app-training-config} records the shared data, optimization, and rollout configuration.
The optimizer is AdamW with a constant learning rate of $1.5\times10^{-6}$ and no warmup.
Each rollout batch contains 1,024 prompts with one response per prompt, followed by minibatches of 256 and one optimization epoch.
Teacher distributions supply the distillation signal; verifier rewards are used for RL and capability evaluation.
Thus, every optimizer step uses the same nominal training budget: 1,024 prompt--response records, four 256-example minibatches, and one optimization pass. This per-step budget is shared by the joint and single-domain runs; the runs differ in domain mixture, not in the number of optimizer steps or the budget assigned to each step.

\begin{table}[htbp]
\centering
\footnotesize
\setlength{\tabcolsep}{3pt}
\caption{Training configuration for Qwen3-4B and Safin1-4B with MOPD methods.}
\label{tab:app-training-config}
\begin{tabular}{@{}p{0.38\textwidth}p{0.27\textwidth}p{0.27\textwidth}@{}}
\toprule
\textbf{Hyperparameter} & \textbf{Qwen3-4B-Instruct} & \textbf{Safin1-4B} \\
\midrule
\multicolumn{3}{l}{\emph{General settings}} \\
Student / teacher architecture & Qwen3-4B-Instruct-2507 & Safin1-4B \\
Configured updates / data seed & 500 / 42 & 500 target / 43 \\
Prompt mixture & 10,000 records per domain & 10,000 records per domain \\
Global rollout batch / optimization minibatch & 1,024 / 256 & 1,024 / 256 \\
Responses per prompt / optimization passes & 1 / 1 & 1 / 1 \\
\midrule
\multicolumn{3}{l}{\emph{Sequence and rollout settings}} \\
Maximum prompt / response length & 16,384 / 16,384 & 16,384 / 32,768 \\
Maximum model length & 32,768 & 49,152 \\
Rollout temperature / top-$p$ & 1.0 / 0.99 & 1.0 / 0.99 \\
Generation top-$k$ & Disabled & Disabled \\
Distillation candidate count & 16 & 16 \\
Thinking mode in the chat template & Enabled & Enabled \\
\midrule
\multicolumn{3}{l}{\emph{Actor optimization}} \\
Optimizer / learning-rate schedule & AdamW / constant & Fused AdamW / constant \\
Learning rate / warmup steps & $1.5\times10^{-6}$ / 0 & $1.5\times10^{-6}$ / 0 \\
Adam momentum coefficients / weight decay & $(0.9,0.999)$ / 0.01 & $(0.9,0.95)$ / 0.10 \\
Gradient norm clip & 1.0 & 1.0 \\
Lower / upper policy clip & 0.20 / 0.28 & 0.20 / 0.28 \\
Dual-clip cap & 3.0 & 3.0 \\
Loss aggregation & Token mean & Token mean \\
Entropy regularization & 0 & Not used \\
Auxiliary actor KL / KL reward shaping & Disabled / disabled & Disabled / disabled \\
\midrule
\multicolumn{3}{l}{\emph{Distributed execution}} \\
Training framework & VERL / FSDP & ms-swift / ZeRO-1 \\
GPUs / vLLM tensor parallelism & 8 H200 / 2 & 8 teacher + 8 student H200 / 2 \\
vLLM memory utilization & 0.34 & 0.60 \\
vLLM max batched tokens & 32,768 & 32,768 \\
Gradient checkpointing / padding removal & Enabled / enabled & Enabled / disabled \\
Actor parameter / optimizer offload & Disabled / disabled & Enabled / enabled \\
\bottomrule
\end{tabular}
\par\smallskip

\end{table}

\paragraph{Method-specific parameters.}
Table~\ref{tab:app-method-config} records the controls that distinguish the three methods.
Open-MOPD combines a target token share with an anchored gap multiplier and normalizes the resulting loss coefficients by their token-weighted mean.
MOPD-Router uses positive cosine alignment, so its operative controls are the candidate count and alignment tolerance.
The learning-rate ablations in Figures~\ref{fig:app-lr-temporal} and~\ref{fig:app-lr-spatial} use $5\times10^{-7}$, $10^{-6}$, $1.5\times10^{-6}$, $2\times10^{-6}$, and $10^{-5}$.

\begin{table}[htbp]
\centering
\small
\setlength{\tabcolsep}{5pt}
\caption{Method-specific settings of MOPD.}
\label{tab:app-method-config}
\begin{tabular}{@{}p{0.20\textwidth}p{0.44\textwidth}p{0.27\textwidth}@{}}
\toprule
\textbf{Method} & \textbf{Control} & \textbf{Value} \\
\midrule
Standard-MOPD & Prompt shares; teacher assignment & $0.2$ each; matched expert \\
Open-MOPD & Target contribution shares & $0.2$ each \\
 & Gap exponent; multiplier clipping & $1.0$; $[0.05,20]$ \\
 & Initial-gap anchoring; advantage refresh & Enabled; enabled \\
MOPD-Router & Teacher aggregation; candidate count & Positive cosine; 16 \\
 & Alignment tolerance & $10^{-6}$ \\
\midrule
Standard / Open & FSDP model storage & FP32 \\
MOPD-Router & FSDP model storage & FP32 \\
\bottomrule
\end{tabular}
\end{table}

\subsection{Prompt Construction and Reward Functions}
\label{app:experiments-prompts}

\paragraph{Prompt construction.}
Examples are stored as role-tagged chat messages and converted with the backbone chat template, preserving task instructions and output delimiters.
The Qwen3 MOPD configuration sets \texttt{enable\_thinking=True}; the domain-specific system messages remain part of the input.
The following instructions describe the saved training records, while evaluation retains the corresponding task-specific answer format.

\paragraph{Math.}
The system instruction is
\begin{quote}\itshape\raggedright
Solve the mathematics problem rigorously. First reason step by step inside \texttt{<think>}\allowbreak\ldots\allowbreak\texttt{</think>}, then put only the final answer in \texttt{\textbackslash boxed\{\}} at the end.
\end{quote}
The user message contains the original mathematical problem.

\paragraph{Code.}
The system message specifies the following reasoning and answer delimiters:
\begin{quote}\itshape\raggedright
The reasoning process and answer are enclosed within \texttt{<think>}\allowbreak\ldots\allowbreak\texttt{</think>} and \texttt{<answer>}\allowbreak\ldots\allowbreak\texttt{</answer>} tags, respectively.
\end{quote}
The user message retains the programming statement and requests a Python code block; input/output examples and execution metadata remain attached to the record.

\paragraph{Science.}
The system instruction is
\begin{quote}\itshape\raggedright
Solve the multiple-choice question carefully. Show your reasoning and put only the final option letter in \texttt{\textbackslash boxed\{\}} at the end.
\end{quote}
Multiple-choice prompts preserve their answer options, while quantitative evaluation questions retain the numerical-answer format.

\paragraph{ToolUse.}
The system message provides the available function names, descriptions, parameter schemas, and call constraints.
Its output instruction requires a call list of the form \texttt{[func1(}\allowbreak\texttt{name=value),} \texttt{func2(}\allowbreak\texttt{name=value)]} with no additional answer text.
The user message supplies the task.
A reasoning instruction precedes the call output, and the verifier parses the final structured call list.

\paragraph{SafeSci.}
The system instruction is
\begin{quote}\itshape\raggedright
You are a precise and knowledgeable assistant. Reason step by step inside \texttt{<think>}\allowbreak\ldots\allowbreak\texttt{</think>} before giving your final answer.
\end{quote}
The user message supplies the scientific question and candidate options and requires a single answer between \texttt{<ans>} and \texttt{<eoa>}.

\paragraph{Reward functions.}
The same domain verifiers provide RL rewards and capability scores.
During MOPD, dense teacher signals determine the optimization target, separating teacher supervision from the verifier outcomes used to evaluate the acquired capabilities.

\paragraph{Math.}
The verifier extracts the final boxed answer and checks mathematical equivalence with the stored answer using the configured symbolic or numerical grader.
Correct answers receive 1; incorrect or unparsable answers receive 0.

\paragraph{Code.}
The generated Python program is executed against the associated input/output or functional tests.
Training retains the fraction of passed tests as partial feedback, while Pass@1 evaluation requires a generated solution to pass all tests.
Execution errors and invalid programs receive no correctness credit.

\paragraph{Science.}
Multiple-choice predictions are checked by exact option matching.
For quantitative SciBench answers, the verifier first compares unit-stripped numerical values with a relative tolerance of $2\%$ and an absolute tolerance of $10^{-6}$, then applies symbolic equivalence when needed.
A verified answer receives 1 and an incorrect answer receives 0.

\paragraph{ToolUse.}
The verifier parses the final call list, normalizes argument values, and checks function names and argument dictionaries against the stored target.
A completely matching call sequence receives 1 and an incorrect or unparsable sequence receives 0.
The separate assistant-text validation records use normalized text matching and do not enter ToolCall accuracy.

\paragraph{SafeSci.}
The verifier extracts the final option from \texttt{<ans>}\ldots\texttt{<eoa>} and checks exact agreement with the stored safe option.
Correct predictions receive 1; incorrect or missing predictions receive 0.

\subsection{Evaluation}
\label{app:experiments-evaluation}

We apply the same benchmark prompts, answer extraction, and verifiers to every method within each backbone block.
Each AIME benchmark contains 30 problems, with 16 responses sampled per problem. We first average binary correctness over the 16 responses to each problem and then average over the 30 problems, yielding Avg@16. Science, ToolUse, and SafeSci use one response per problem; Code reports Pass@1 from one response per problem.
Response caps follow Table~\ref{tab:app-models}.
Scores are first averaged within each benchmark, then within each domain, and finally across the five domains, following Section~\ref{sec:temporal settings}.
Writing $c_{k,u}$ for the percentage score of benchmark $u$ in domain $k$ and $\mathcal B_k$ for the associated benchmark set, the domain macro-average is
\begin{equation}
\label{eq:app-macro-average}
C_{\mathrm{macro}}=\frac{1}{5}\sum_{k=1}^{5}\left(\frac{1}{|\mathcal B_k|}\sum_{u\in\mathcal B_k}c_{k,u}\right).
\end{equation}
The domain curves in both panels of Figure~\ref{fig:T.3} average the three Math, two Science, and two Code benchmarks within their respective domains.

\subsection{Checkpoint and Probe Protocol}
\label{app:experiments-anchors}

We trace 39 nonuniform joint-training anchors: steps $0$, $1$, $2$, and $5$; every tenth step from $10$ to $200$; every twentieth step from $220$ to $500$.
The main joint-training traces span 500 actual optimizer steps.
The intervention curves use actual optimizer steps throughout.
Geometric trajectories use $\log(1+t/10)$ spacing, whereas capability trajectories and the coordination diagnostics below use linear training steps.
The spatial reference figures compare current subspaces with previous and initial anchors, as defined in Appendix~\ref{app:reference-geometry}.
The overview and learning-rate figures retain the original same-time cross-domain readout.
Appendix~\ref{app:tracing-protocol} details the spectral and subspace estimators.

\paragraph{Probe population and frozen inputs.}
The geometry code reads the \texttt{prompt} column of the five-domain training parquet, selecting $M=10{,}000$ records per domain without replacement using seed 0. Thus Math, Science, Code, ToolUse, and SafeSci each contribute 10,000 rows.
The fixed objects are the complete formatted and truncated prompt token contexts, including system instructions and template tokens. The per-domain hashes of the ordered formatted inputs and all recorded tokenizer, truncation, and extraction settings agree across the 42 anchors of each of the three MOPD methods.


\subsection{Domain-Intervention Protocol and Final Evaluation}
\label{app:domain-interventions}

Figures~\ref{fig:T.4} and~\ref{fig:S.3} compare the five-domain Standard-MOPD baseline with five four-domain mixtures.
Each intervention removes all 10,000 training records of one domain and its matched teacher from hard-routed distillation.
The remaining 40,000 records retain their original prompts and matched experts.
There is one training trajectory per arm.
Equal retained dataset sizes change expected prompt shares from $1/5$ to $1/4$; fixing steps and batch size controls the total prompt budget while increasing each retained domain's expected exposure by $25\%$.

All six trajectories complete 500 optimizer steps and final validation at step 500.
Each curve contains 42 measured anchors from initialization through step 500.
The main-text numerical intervention comparisons use the raw step-500 geometric measurements and the corresponding final validation scores.
All runs are evaluated geometrically on the same five frozen 10,000-record probe populations.
The saved ordered input hashes and extraction configuration agree across arms.
Features are the last nonpadding prompt-token hidden states from the final transformer block (layer 36), using a maximum prompt length of 512 tokens.
Centered covariance spectra use relative eigenvalue floor $10^{-12}$, and spatial comparisons use the leading 16 eigenvectors.
Cross-domain averages always include the other four probe domains within the same model, regardless of their presence in training.

Table~\ref{tab:domain-intervention-results} reports final evaluations of the six step-500 checkpoints, together with the initial student, using the suite in Appendix~\ref{app:experiments-evaluation}.
Math averages AIME24, AIME25, and AIME26 Avg@16; Science averages SciBench and GPQA; Code averages LiveCodeBench v5 and v6.
ToolUse reports ToolCall accuracy, excluding the assistant-text diagnostic, and SafeSci uses its domain benchmark accuracy.
Overall gives each domain equal weight: we first average benchmarks within each domain and then average the five domain scores, without intermediate rounding.
For this table, aggregation uses the reported two-decimal benchmark percentages and rounds Overall to two decimals.
The fixed-step intervention table is distinct from the method-comparison summary in Table~\ref{tab:qwen3-five-domain-results}.

\subsection{Learning-Rate Sensitivity}
\label{app:learning-rate}

\begin{figure*}[htbp]
  \centering
  \includegraphics[width=\textwidth]{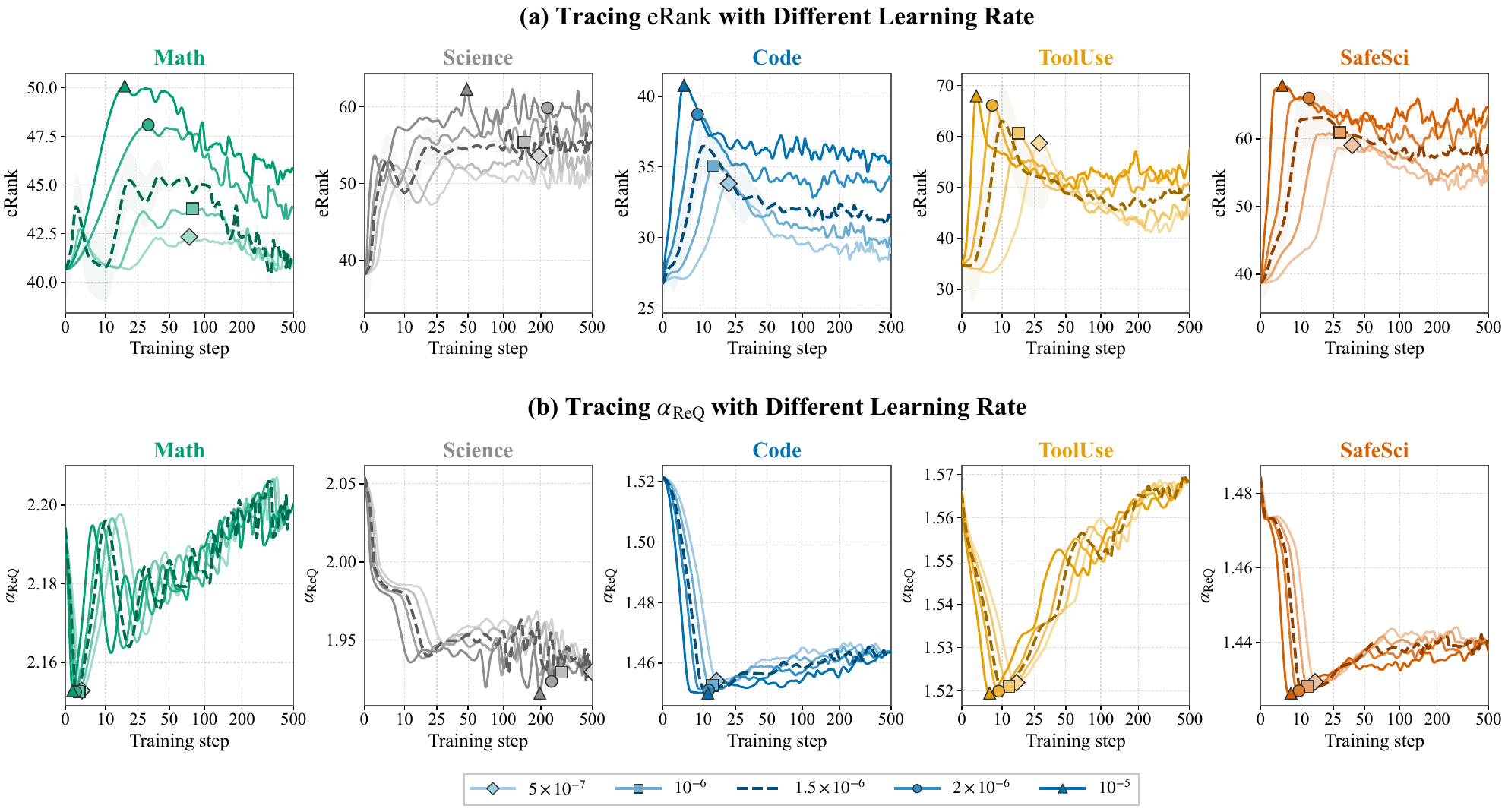}
  \caption{\textbf{Learning-rate sensitivity of temporal geometry.}
  Five-domain Standard-MOPD trajectories show (a) $\mathrm{eRank}$ and (b) $\alpha_{\mathrm{ReQ}}$; outlined markers indicate maxima and minima, respectively.
  The default rate is $1.5\times10^{-6}$.}
  \label{fig:app-lr-temporal}
\end{figure*}

\begin{figure*}[htbp]
  \centering
  \includegraphics[width=\textwidth]{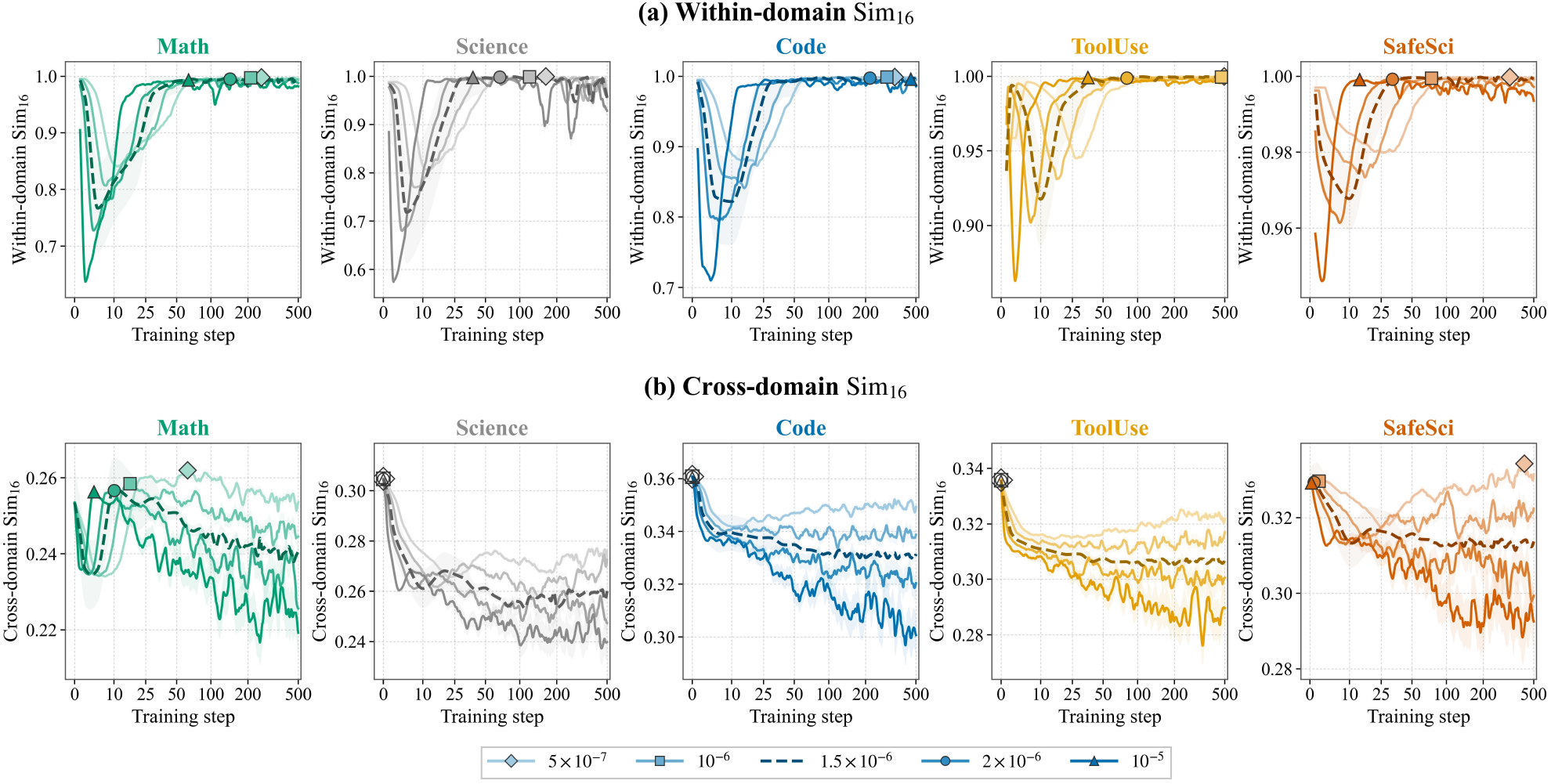}
  \caption{\textbf{Learning-rate sensitivity of spatial geometry.}
  (a) Within-domain $\operatorname{Sim}_{16}$ between consecutive saved anchors; (b) mean cross-domain $\operatorname{Sim}_{16}$ at the same anchor.
  Learning rates match Figure~\ref{fig:app-lr-temporal}.}
  \label{fig:app-lr-spatial}
\end{figure*}

We vary the constant learning rate of Standard-MOPD over
$\{5\times10^{-7},10^{-6},1.5\times10^{-6},2\times10^{-6},10^{-5}\}$,
with $1.5\times10^{-6}$ as the default.
Figure~\ref{fig:app-lr-temporal} traces spectral responses, and Figure~\ref{fig:app-lr-spatial} traces adjacent-anchor within-domain and same-time cross-domain overlap.
These figures retain the original learning-rate measurements and rendering.
Horizontal spacing follows $\log(1+t/10)$, with tick labels giving training steps.
Outlined temporal markers identify displayed $\mathrm{eRank}$ maxima and $\alpha_{\mathrm{ReQ}}$ minima.

The update scale changes the timing and magnitude of early expansion, the subsequent contraction, and later spectral variability.
Larger updates generally produce earlier pronounced changes, while the highest rate introduces greater later fluctuation in several domains.
The extent and timing of these responses differ by domain; changing the rate does not synchronize all five trajectories.
Across the displayed rates, the spectral curves retain the broad expansion--consolidation--slower-change pattern motivating ECO, with substantial differences in peak scale and late level.

The spatial traces show a corresponding change in the pace of reorganization.
Their same-time cross-domain readout differs from the previous- and initial-reference comparisons in Figures~\ref{fig:S.2} and~\ref{fig:S.3}.
Higher rates generally advance and deepen the early adjacent-anchor similarity drop, followed by recovery toward high local persistence.
Cross-domain overlap differentiates earlier at larger rates, with domain-dependent later levels.
Thus, a similar later degree of within-domain stability can follow different early paths and coexist with different cross-domain organization.

\subsection{Supervision Coordination across Five Domains}
\label{app:experiments-coordination}

The geometric traces describe the internal response to training, while the coordination quantities expose the external coefficients that generate the update in Eq.~\eqref{eq:update}.
We inspect all 500 logged training steps of Open-MOPD and MOPD-Router, preserving the global step when combining resumed training segments.
The following diagnostics connect domain exposure and teacher allocation to the heterogeneous adaptation discussed in Sections~\ref{sec:efficiency},~\ref{sec:within}, and~\ref{sec:across}.

\begin{figure*}[htbp]
\centering
\includegraphics[width=\textwidth]{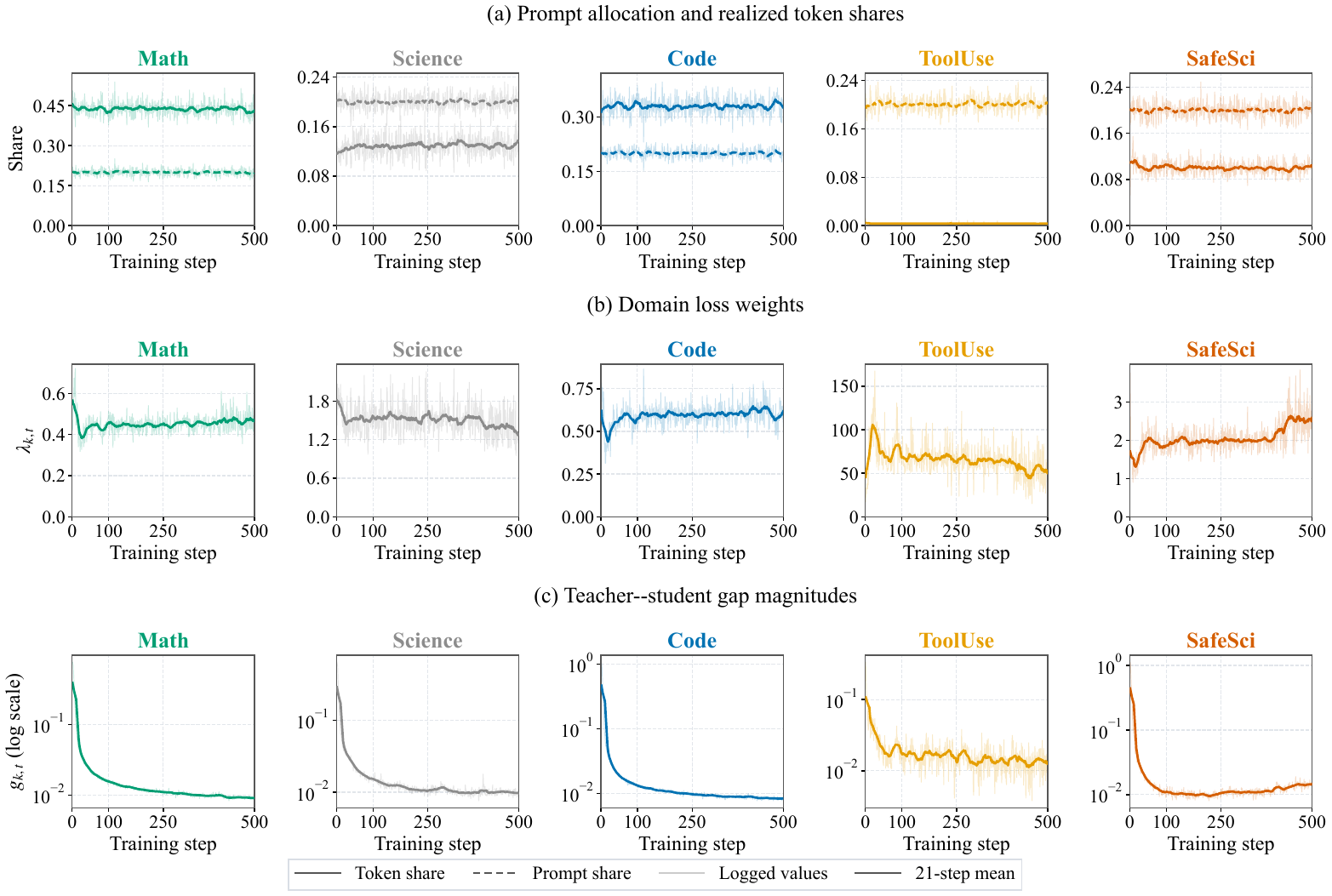}
\caption{\textbf{Five-domain supervision coordination in Open-MOPD.}
(a) Realized token shares (solid) and prompt shares (dashed); (b) normalized domain loss coefficients; (c) mean absolute distillation signals on logarithmic vertical axes.}
\label{fig:app-open-coordination}
\end{figure*}

\paragraph{Open-MOPD: effective domain allocation.}
Let $\nu_{k,t}$ be the observed response-token share, $p_k^\star=0.2$ the target contribution, and $g_{k,t}$ the mean absolute token-level distillation signal of domain $k$.
With $g_{k,\mathrm{init}}$ the initial gap statistic of the active training segment, the saved loss coefficient is
\begin{equation}
\label{eq:app-open-weight}
\lambda_{k,t}=\frac{\displaystyle\frac{p_k^\star}{\nu_{k,t}}\operatorname{clip}\!\left(\frac{g_{k,t}}{g_{k,\mathrm{init}}},0.05,20\right)}{\displaystyle\sum_{d=1}^{K}\nu_{d,t}\frac{p_d^\star}{\nu_{d,t}}\operatorname{clip}\!\left(\frac{g_{d,t}}{g_{d,\mathrm{init}}},0.05,20\right)}.
\end{equation}
This normalization preserves a token-weighted mean coefficient of one while reallocating loss contributions.
Figure~\ref{fig:app-open-coordination} separates prompt shares, realized token shares, loss weights, and gap magnitudes.
Over steps 400--500, ToolUse contributes about $20.04\%$ of prompts but only $0.33\%$ of response tokens, accompanied by a mean loss weight of $54.90$.
Math contributes $43.71\%$ of tokens and receives a mean coefficient of $0.47$.
Thus, balanced prompt counts coexist with sharply unequal token exposure, and temporal coordination compensates for this imbalance without changing the prompt mixture.
The declining gap curves further show why a fixed initial correction would not track the evolving distillation signal.

\begin{figure*}[htbp]
\centering
\includegraphics[width=\textwidth]{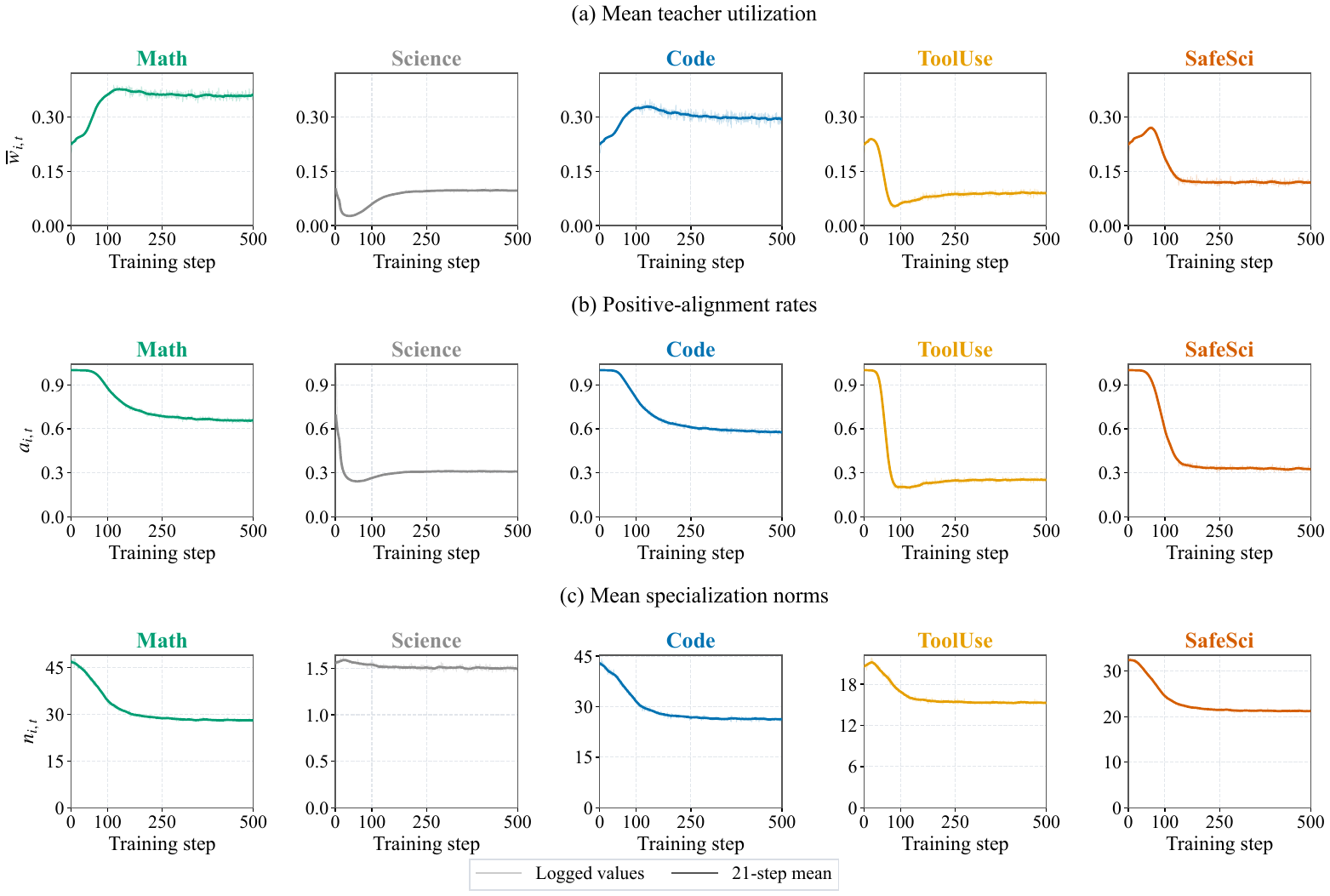}
\caption{\textbf{Five-expert coordination in MOPD-Router.}
(a) Mean teacher weight; (b) eligible positive-alignment rate; (c) mean specialization norm on the student top-16 candidate set.
Columns denote teachers; statistics aggregate valid tokens across all prompt domains.}
\label{fig:app-router-coordination}
\end{figure*}

\paragraph{MOPD-Router: teacher allocation and directional selection.}
At each student-visited state $s$, the router compares teacher specialization from the common initial checkpoint with the teaching direction from the current student.
Let $\chi_{i,t}(s)$ indicate a positive alignment passing the configured tolerance, and let $\delta_i(s)$ denote the specialization vector on the top-16 candidate set.
Averaging over the valid response tokens $\mathcal S_t$ gives the three logged quantities
\begin{equation}
\label{eq:app-router-diagnostics}
\big(\overline w_{i,t},a_{i,t},n_{i,t}\big)=\frac{1}{|\mathcal S_t|}\sum_{s\in\mathcal S_t}\big(w_{i,t}(s),\chi_{i,t}(s),\|\delta_i(s)\|_2\big).
\end{equation}
Each column in Figure~\ref{fig:app-router-coordination} identifies an expert domain; its statistics aggregate tokens from all five prompt domains.
During steps 400--500, the Math and Code experts retain mean utilization weights of $0.358$ and $0.296$, whereas the other experts each receive approximately $0.09$--$0.12$.
Alignment rates and specialization norms also differ across experts, showing that a joint teacher pool develops an unequal pattern of usable directional supervision.
These changing weights clarify how $w_{i,t}(x,y)$ redistributes external signals, while the overlaps in Figure~\ref{fig:S.1} describe the internal organization.

\clearpage
\section{More Algorithmic Analyses}
\label{app:tracing-protocol}

\subsection{Spectral Estimation and Effective Dimension}
\label{app:estimation}

For $M$ centered $d$-dimensional probes, $\operatorname{rank}(\Sigma_{k,t})\leq\min(d,M-1)$. If $\Sigma_{k,t}\neq0$, the effective rank in Eq.~\eqref{eq:erank} is between one and this algebraic rank. It equals $r$ when the $r$ positive eigenvalues are equal and equals one when all variance lies in one direction. Normalizing the eigenvalues removes the overall covariance scale: a change in eRank reflects a change in spectral shape, not necessarily a larger or smaller feature norm. The zero matrix has no normalized spectrum and should be marked unavailable.
The $\alpha_{\mathrm{ReQ}}$ fit in Eq.~\eqref{eq:areq} uses eigenvalues greater than $10^{-12}\sigma_1$ and ordinary least squares in log-rank/log-eigenvalue coordinates.

\subsection{Principal Subspaces and Their Comparison}
\label{app:subspaces}

For orthonormal $d\times K_s$ bases $Q$ and $Q'$ with $K_s=16$, the singular values of $Q^\top Q'$ are cosines of the principal angles. Consequently,
\begin{equation}
\label{eq:app-principal-overlap}
\operatorname{Sim}_{16}(Q,Q')=\frac{\|Q^\top Q'\|_F^2}{16}
=\frac{1}{16}\sum_{a=1}^{16}\cos^2\phi_a.
\end{equation}
The score is unchanged by rotating either basis within its span. Coincident spans score one; orthogonal spans score zero when the ambient dimension allows them. A small eigengap around the sixteenth eigenvalue can make the estimated top-16 span sensitive to perturbations, even if the overall covariance changes little. The numerical covariance rank must also be at least 16.

Within-domain comparisons of checkpoints from one model use a common feature coordinate system. Cross-domain comparisons at the same or different checkpoints within that model do so as well. Independently trained models may differ by feature rotations or other reparameterizations, so direct cross-model subspace overlap requires explicit alignment. We therefore compare single-domain and multi-domain trajectories through their own within-model overlaps or scalar spectral summaries.

\subsection{Previous- and Initial-Reference Geometry}
\label{app:reference-geometry}

Let $t_{\mathrm{prev}}$ be the previous saved anchor in the same run and let $r\in\{t_{\mathrm{prev}},0\}$ select the temporal reference.
With the domain--time notation of Eq.~\eqref{eq:simk}, define
\begin{equation}
\label{eq:spatial-reference-comparisons}
\begin{aligned}
W_{k,t}^{[r]}&=\operatorname{Sim}_{16}\big((k,t),(k,r)\big),\\
C_{k,t}^{[r]}&=\frac{1}{K-1}\sum_{j\neq k}
\operatorname{Sim}_{16}\big((k,t),(j,r)\big),\qquad K=5.
\end{aligned}
\end{equation}
All bases in either expression come from one model trajectory in its common hidden-state coordinate system.
$W^{[t_{\mathrm{prev}}]}$ measures local persistence; $W^{[0]}$ measures overlap with the domain's original span.
$C^{[t_{\mathrm{prev}}]}$ measures overlap with the other domains' recent representations, while $C^{[0]}$ uses their fixed initial representations.
Because the recent reference also evolves, a change in $C^{[t_{\mathrm{prev}}]}$ describes a changing relation between subspaces rather than movement toward a fixed target.
For the overview and learning-rate plots only, $C_{k,t}^{[t]}$ gives the original same-time cross-domain readout.
In intervention runs, $K=5$ still counts all frozen probe domains, including the omitted training domain.

At $t=0$, no previous saved anchor exists.
The plotted boundary is $W_{k,0}=1$ and the shared initial cross-domain overlap for $C_{k,0}$, not a measured pre-initialization transition.
At later anchors, the previous-reference gap is the actual saved-step interval, which increases from 10 to 20 steps after step 200 in the intervention runs.
Initial-reference comparisons are unaffected by this change in the adjacent-anchor interval.

\subsection{Numerical Diagnostics and Subspace-Dimension Sensitivity}
\label{app:geometry-robustness}

We audit the saved Qwen3 geometry records for Standard-MOPD, Open-MOPD, and MOPD-Router: 42 anchors and five domains per method, totaling 630 domain covariance estimates. Each estimate averages eight sampling repetitions over $M=10{,}000$ fixed prompts from the training parquet, with feature width $d=2560$, following Appendix~\ref{app:experiments-anchors}. The ordered probe-input hashes and recorded extraction settings agree across all 126 method--checkpoint records, fixing the input population as the student changes.
All 630 covariances have numerical rank 2560 at the relative floor $10^{-12}$, and every $\alpha_{\mathrm{ReQ}}$ regression retains all 2560 eigenvalues. The log--log fit $R^2$ ranges from $0.8777$ to $0.9427$. The relative boundary eigengap $(\sigma_{16}-\sigma_{17})/\sigma_{16}$ ranges from $0.0050$ to $0.2170$.

We recompute overlaps at $K_s\in\{8,12,16\}$ using the corresponding leading columns of the saved bases. Recomputing $\operatorname{Sim}_{16}$ reproduces the saved within-domain and cross-domain values to a maximum absolute difference below $10^{-12}$.
Table~\ref{tab:app-subspace-sensitivity} summarizes the later regime, defined here as anchors at $t\geq200$. 
Within-domain entries collect the consecutive-anchor comparisons. 
Cross-domain entries give the range of each domain's mean overlap with the other four domains across those anchors.

\begin{table}[t]
\centering
\small
\setlength{\tabcolsep}{5pt}
\caption{Subspace-dimension sensitivity on fixed Qwen3 probes at $t\geq200$.}
\label{tab:app-subspace-sensitivity}
\begin{tabular}{@{}lrrrr@{}}
\toprule
Method & $K_s$ & Within min. & Within median & Cross-domain range \\
\midrule
Standard-MOPD & 8  & 0.9818 & 0.9972 & $[0.2002,0.3685]$ \\
              & 12 & 0.9664 & 0.9965 & $[0.2417,0.3532]$ \\
              & 16 & 0.9469 & 0.9966 & $[0.2378,0.3322]$ \\
\midrule
Open-MOPD     & 8  & 0.9850 & 0.9976 & $[0.2006,0.3768]$ \\
              & 12 & 0.9706 & 0.9968 & $[0.2442,0.3580]$ \\
              & 16 & 0.9585 & 0.9962 & $[0.2420,0.3339]$ \\
\midrule
MOPD-Router   & 8  & 0.9966 & 0.9995 & $[0.2590,0.3899]$ \\
              & 12 & 0.9985 & 0.9995 & $[0.2573,0.3703]$ \\
              & 16 & 0.9933 & 0.9993 & $[0.2508,0.3425]$ \\
\bottomrule
\end{tabular}
\end{table}

High later within-domain persistence and partial cross-domain overlap hold at all three dimensions. The magnitudes and early minima depend on $K_s$: for example, MOPD-Router's minimum consecutive-anchor overlap over the full trajectory is $0.9747$, $0.9122$, and $0.9533$ at $K_s=8,12,16$, respectively. The main-text threshold of $0.95$ therefore refers specifically to $K_s=16$.
These checks assess numerical consistency and dimension sensitivity on the observed trajectories.

\end{document}